\documentclass[times, review, 10pt]{elsarticle}

\usepackage{amssymb}
\usepackage{amsmath}

\usepackage{graphicx}%
\usepackage{multirow}%
\usepackage{amsfonts}%
\usepackage{amsthm}%
\usepackage{mathrsfs}%
\usepackage[title]{appendix}%
\usepackage{xcolor}%
\usepackage{textcomp}%
\usepackage{manyfoot}%
\usepackage{booktabs}%
\usepackage{algorithm}%
\usepackage{algorithmicx}%
\usepackage{algpseudocode}%
\usepackage{listings}%

\usepackage{hyperref}
\usepackage{url}
\usepackage{mathtools}
\usepackage{multirow}
\usepackage{makecell}
\usepackage{wrapfig}
\usepackage{siunitx}
\usepackage{color}
\usepackage{setspace}
\usepackage{amsthm}
\usepackage{ragged2e}
\usepackage{subcaption}

\definecolor{red}{rgb}{0,0,0}

\newcommand{\valpha}{\boldsymbol{\alpha}}
\newcommand{\st}{\mathrm{s.t.}}
\newcommand{\diag}{\mathrm{diag}}
\newcommand{\vh}{\boldsymbol{h}}
\newcommand{\vd}{\boldsymbol{d}}
\newcommand{\vu}{\boldsymbol{u}}
\newcommand{\vv}{\boldsymbol{v}}
\newcommand{\vx}{\boldsymbol{x}}
\newcommand{\vo}{\boldsymbol{o}}
\newcommand{\vw}{\boldsymbol{w}}
\newcommand{\vy}{\boldsymbol{y}}
\newcommand{\vmu}{\boldsymbol{\mu}}

\newcommand{\mA}{\boldsymbol{A}}
\newcommand{\mK}{\boldsymbol{K}}
\newcommand{\mG}{\boldsymbol{G}}
\newcommand{\mM}{\boldsymbol{M}}
\newcommand{\mB}{\boldsymbol{B}}
\newcommand{\mI}{\boldsymbol{I}}
\newcommand{\mX}{\boldsymbol{X}}
\newcommand{\mU}{\boldsymbol{U}}
\newcommand{\mV}{\boldsymbol{V}}
\newcommand{\mO}{\boldsymbol{O}}
\newcommand{\mY}{\boldsymbol{Y}}
\newcommand{\mSigma}{\boldsymbol{\Sigma}}

\newcommand{\vzero}{\boldsymbol{0}}
\newcommand{\vone}{\boldsymbol{1}}

\newcommand{\argmin}{\mathop{\arg\min}}

\theoremstyle{thmstyleone}%
\newtheorem{theorem}{Theorem}
\newtheorem{lemma}[theorem]{Lemma}
\newtheorem{assumption}{Assumption}

\newtheorem{property}[theorem]{Property}

\theoremstyle{thmstyletwo}%
\newtheorem{remark}{Remark}%

\theoremstyle{thmstylethree}%
\journal{Pattern Recognition}

\begin{document}




\begin{frontmatter}



\title{Data Imputation by Pursuing Better Classification: A Supervised Kernel-Based Method}


\author[sjtu]{Ruikai Yang} 
\ead{ruikai.yang@sjtu.edu.cn}

\author[sjtu]{Fan He} 
\ead{hf-inspire@sjtu.edu.cn}
\author[sjtu]{Mingzhen He} 
\ead{mingzhen\_he@sjtu.edu.cn}
\author[sjtu]{Kaijie Wang} 
\ead{kaijie\_wang@sjtu.edu.cn}
\author[sjtu]{Xiaolin Huang\corref{cor1}} 
\ead{xiaolinhuang@sjtu.edu.cn}

\cortext[cor1]{Corresponding author.}

\affiliation[sjtu]{organization={Institute of Image Processing and Pattern Recognition, Department of Automation, Shanghai Jiao Tong University},
            city={Shanghai},
            postcode={200240}, 
            country={China}}


\begin{abstract}
Data imputation, the process of filling in missing feature elements for incomplete datasets, plays a crucial role in data-driven learning. A fundamental belief is that data imputation is helpful for learning performance, and it follows that the pursuit of better classification can guide the data imputation process. While some works consider using label information to assist in this task, their simplistic utilization of labels lacks flexibility and may rely on strict assumptions. In this paper, we propose a new framework that effectively leverages supervision information to complete missing data in a manner conducive to classification. Specifically, this framework operates in two stages. Firstly, it leverages labels to supervise the optimization of similarity relationships among data, represented by the kernel matrix, with the goal of enhancing classification accuracy. To mitigate overfitting that may occur during this process, a perturbation variable is introduced to improve the robustness of the framework. Secondly, the learned kernel matrix serves as additional supervision information to guide data imputation through regression, utilizing the block coordinate descent method. The superiority of the proposed method is evaluated on four real-world datasets by comparing it with state-of-the-art imputation methods. Remarkably, our algorithm significantly outperforms other methods when the data is missing more than 60\% of the features.
\end{abstract}




\end{frontmatter}



\section{Introduction}
\label{sec1}

The presence of missing data poses significant challenges in machine learning and data analysis~\cite{little2019statistical, su2024nonmonotone,mao2025robust}. In many real-world application scenarios, ensuring that all entries in the data are complete is a difficult task, and this can be attributed to various reasons. Firstly, it may occur due to limitations in data collection, such as time constraints or limited resources, resulting in certain observations being omitted. Secondly, data can be intentionally designed to have missing values, for instance, in survey studies where participants may choose not to answer certain questions, thereby introducing missing values into the dataset. By completing missing data, researchers can uncover the underlying structure and relationships within the dataset, optimizing the utilization of the complete dataset to enhance the performance and efficiency of subsequent analyses. Therefore, the significance of filling in missing data becomes self-evident.

\begin{figure}[tbp]
    \centering
    \includegraphics[width=0.65\linewidth]{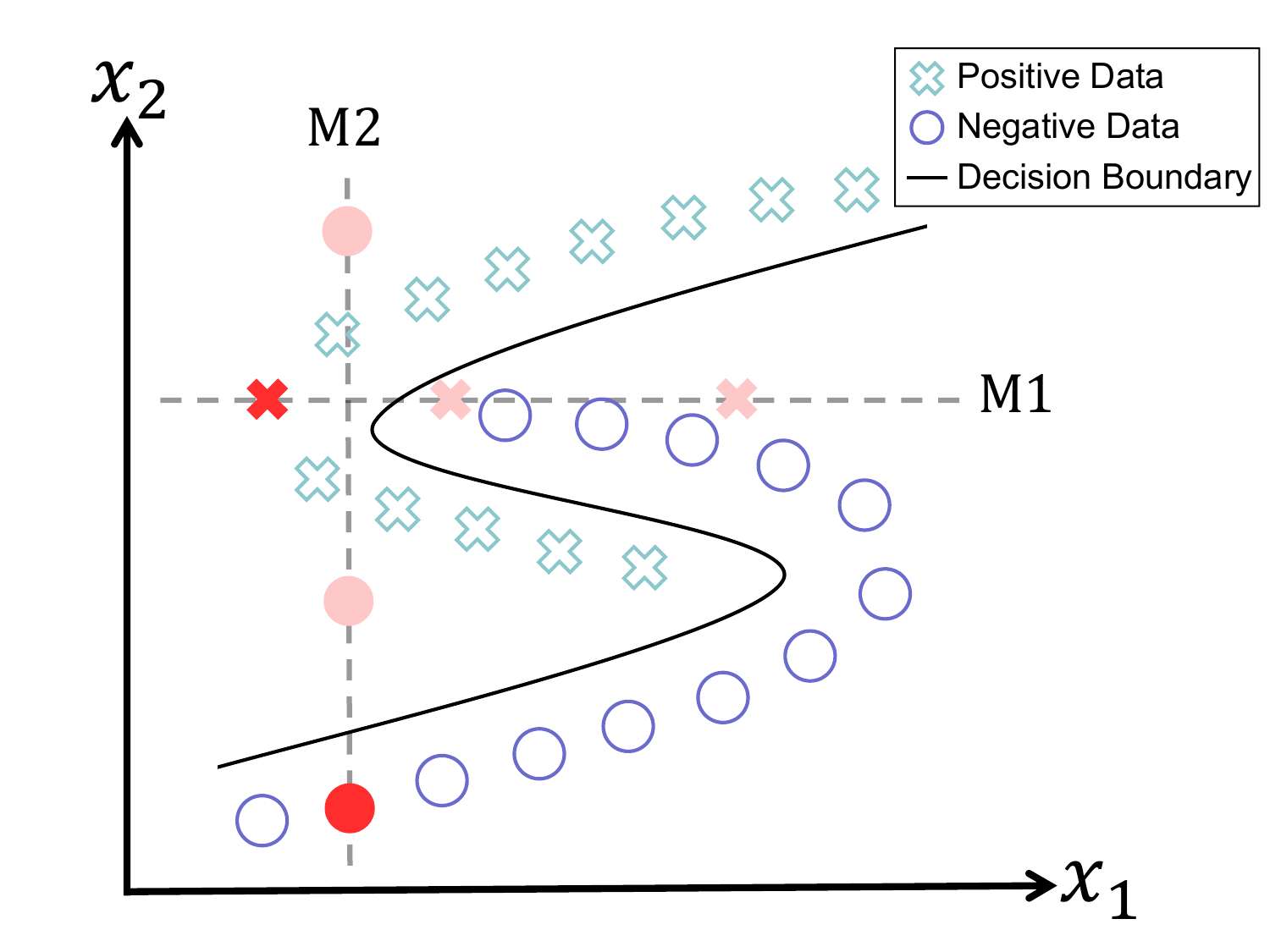}
	\caption{Imputation of the two moons dataset. Positive and negative data are denoted by X and O, while the solid curve denotes the ideal decision boundary of the classifier. The gray dashed lines M1 and M2 represent all possible imputations for a positive data with a missing $x_1$ dimension and a negative data with a missing $x_2$ dimension, respectively. Among them, the red results are more likely to lead to better outcomes in the subsequent classification task compared to the pink results, making them more desirable.}
    \label{Fig: synthetic}
\end{figure}

The existing data imputation methods primarily focus on the relationships among features. For example, mean imputation (MI, \citep{schafer1997analysis}) considers the mean value of the feature, while other methods consider the low-rank property of the imputed matrix~\citep{candes2010power, cai2013max, xu2020distributed} or incorporate similarity information~\citep{batista2002study, smieja2019generalized}. \textcolor{black}{Some methods also utilize label information, with some performing different operations on data from different classes~\citep{smola2005kernel, allison2009missing}, while others treat labels as additional features and then complete the matrix~\citep{goldberg2010transduction}. However, these utilizations of label information in imputation models lack flexibility and are insufficient to depict the complex relationship between data and labels.} Considering a fundamental observation that \emph{data are helpful for distinguishing the labels}, we can derive an imputation criterion: 
\[
\emph{data imputation should lead to improved classification performance.}
\]
Thus we expect imputation results that can perform better on subsequent tasks. Let us consider a toy example illustrated in Figure~\ref{Fig: synthetic}. In the synthetic two moons dataset, we have a positive data with a missing value in the $x_1$ dimension and a negative data with a missing value in the $x_2$ dimension. All possible imputations for these two data are represented by gray dashed lines, labeled as M1 and M2. By applying the proposed imputation criterion, we prioritize selecting the red results over the pink results to improve the subsequent classification task.

In a given sequence of triplets $\{(\vx_i, \vo_i, y_i)\}_{i=1}^N$, where $\vx_i\in\mathbb{R}^d$ and $y_i\in\mathbb{R}$, we typically use the set of observable features $\vo_i\subseteq 2^{\lbrace 1,\cdots,d\rbrace }$ and missing features $\ast$ to represent the missing data $\vx_{\vo_i}^i\in(\mathbb{R}\cup \{\ast\})^d$. This sequence is then consolidated into an incomplete dataset $\mathcal{D}=\lbrace \vx_{\vo_i}^i, y_i\rbrace _{i=1}^N$. The task now becomes to find suitable missing values that aid in classification. For kernel-based methods widely applied in machine learning, such as support vector machine (SVM,~\citep{scholkopf2002learning, steinwart2008support, wang2023consensus}), a significant characteristic is that classifiers can be explicitly written as functions of the relationship between data and labels. The relationship between data will be defined by a kernel function, and different kernel functions provide distinct measures of similarity. Specifically, a kernel matrix or Gram matrix, denoted as $\mK \in \mathbb{R}^{N\times N}$, is constructed, where $K_{i,j} = k(\vx_i,\vx_j)$ represents the similarity between data $\vx_i$ and $\vx_j$ using a proper positive definite kernel function $k(\cdot,\cdot):\mathbb{R}^d\times\mathbb{R}^d\rightarrow\mathbb{R}$. A classifier is then trained based on this kernel matrix $\mK$. When handling missing data, represented as $\vx_{\vo_i}^i$, the value of $K_{i,\cdot}\in\mathbb{R}^N$ is unknown. Thus, it is necessary to find $\tilde{K}_{i,j}$ for data with missing features in order to establish appropriate relationships with other data. In SVM, this process can be seamlessly integrated with supervised information, leading to enhanced prediction accuracy in subsequent classification tasks. By considering prior knowledge of missing values and imposing constraints on the adjustment of the kernel matrix, we can effectively incorporate this information into the modeling process. 
The next step is to recover the missing values from the imputed matrix $\tilde{\mK}$. This step is relatively easy, particularly in the traditional scenario where $N \gg d$. Because for each data, there is $N - 1$ available supervised information stored in the kernel matrix to guide the imputation of $d$ or fewer elements. In this paper, we propose a two-stage framework for data imputation: I) the kernel matrix is completed by pursuing high classification accuracy; II) the missing features are reconstructed to approximate the optimized kernel matrix. With the proposed method, we can find good estimates for missing values that lead to better classification performance in the subsequent task. Therefore, when faced with situations depicted in Figure \ref{Fig: synthetic}, we are able to obtain the imputation results indicated by the red crosses and circles.

In the first stage, we begin by computing the initial kernel matrix based on the observed features. This matrix is then further modified through a process referred to as \textit{kernel matrix imputation} throughout this paper. Recent studies~\citep{liu2020learning, wang2023learning} have demonstrated that integrating kernel learning and classifier optimization tasks, and alternating between them, can improve classification performance and naturally lead to a fine-tuned kernel matrix. The effectiveness of such joint learning ideas has recently been also confirmed in neural network-based methods as well~\citep{le2021sa}. Inspired by these ideas, we adopt a similar approach by combining the tasks of kernel matrix imputation and classifier training into a maximum-minimum optimization problem. However, for such joint optimization objectives, we need to ensure that the kernel matrix is optimized to achieve high classification accuracy while also making the classifier sufficiently robust. Otherwise, the flexibility of modifying the similarity relationships between data can make the classifier susceptible to overfitting. To prevent this situation, we introduce an additional perturbation matrix during the optimization of the kernel matrix.
In the second stage, we utilize the block coordinate descent (BCD,~\citep{tseng2001convergence}) method to solve a non-convex problem. Although finding the global optimal solution is challenging, our numerical experiments demonstrate that when the data size is much larger than the feature dimension, the abundant supervisory information from the kernel matrix effectively guides the imputation process. As a result, we can still achieve highly accurate solutions. 
The main contributions of this paper are summarized as follows:
\begin{itemize}
    \item To fully leverage label information, we propose a novel two-stage imputation framework. This framework optimizes the similarity relationships between data by utilizing supervision information and guides the imputation accordingly.
    \item We have developed a nonparametric method to impute the kernel matrix, which is performed alternately with the training of the classifier. By introducing a perturbation variable, we enhance the robustness of the classifier.
    \item We provide a solving algorithm based on the BCD to accurately recover missing data from a given kernel matrix. This algorithm effectively utilizes the learned similarity information among the training data.
    \item Experimental results on real datasets indicate that the imputed data generated by our framework exhibits excellent performance in classification tasks. Notably, when the missing ratio exceeds 60\%, our algorithm outperforms other imputation methods significantly.
\end{itemize}

\section{Related Work}\label{sec2}

The task of learning a classifier from missing data has been extensively studied in the past few decades, leading to the development of three main approaches as below.

\textbf{Filling data before classification.} A large portion of the research focuses on filling in the data before using it in the subsequent classification task. For instance, a framework based on maximum likelihood density estimation for coping with missing data was introduced, utilizing mixture models and the expectation-maximization principle~\citep{ghahramani1993supervised}. A method that modeled the missing values as Gaussian random variables and estimated their mean and covariance from the observed data and labels, which were then used in a modified SVM, was proposed~\citep{bhattacharyya2004second}. Later, a Gaussian mixture model was employed to estimate the conditional probability of missing values given the observed data and impute them using expectations~\citep{williams2007classification}. Additionally, some work estimated complete data by computing the corresponding marginal distributions of the missing data~\citep{smola2005kernel} or assumed a low-rank subspace for the data and treated labels as an additional feature to handle missing values~\citep{goldberg2010transduction}. In recent years, a kernel-based multi-scale data imputation method was proposed~\citep{rabin2020multi}, and the known matrix structure in terms of subspaces was utilized to assist in matrix imputation~\citep{lopez2023near}. In addition, there are currently many deep learning-based imputation methods, such as those based on autoencoders~\cite{mattei2019miwae}, generative adversarial networks~\cite{yoon2018gain,qin2024improved}, attention-based approaches~\cite{kowsar2024attention}, and methods leveraging diffusion models~\cite{chen2024rethinking,lu2025diffusion}. But these methods separate the imputation process from the classification task, which may not necessarily lead to improved performance with their imputation outcomes.

\textbf{Filling the kernel matrix before classification.} Another research direction focuses on kernel-based machine learning algorithms. In this approach, classification or regression tasks only require the kernel matrix based on the training data. Therefore, methods have been explored to compute kernel function values between data with missing values instead of directly imputing the missing data. A parameterized kernel function based on the low-rank assumption of data was proposed, which allows for input vectors with missing values~\citep{hazan2015classification}. However, when computing similarities, only the dimensions present in both data are considered, and the rest are discarded. There is also work that modeled the squared distance between two missing data as a random variable following a gamma distribution and computed the expectation of the Gaussian kernel function under missing data~\citep{mesquita2019gaussian}. Similarly, possible outcomes of missing values were modeled with the data distribution, and the expectation of similarity between two missing data was computed~\citep{smieja2019generalized}. Still, these methods have limitations in flexibly utilizing the supervision information from labels during the imputation process, which consequently restricts their performance in practical tasks.

\textbf{Direct classification of incomplete data.} Researchers have also explored methods for directly classifying missing data. In one study, a modified risk was defined to address the uncertainty in prediction results caused by missing data~\citep{pelckmans2005handling}. Another approach maximized the margin of the subspace correlated with the observed data~\citep{chechik2008max}. However, when computing kernel values, they encounter similar issues of incomplete information utilization as in~\citep{hazan2015classification}. In a different approach, missing values were treated as a specific type of noise, and a linear programming problem resembling SVM formulation was developed to learn a robust binary classifier~\citep{dekel2010learning}. And some works optimize the nonlinear classifier while seeking the linear subspace where the data might lie~\citep{sheikholesalmi2014classification, xu2020distributed}. Due to the necessity of predefining the dimension of the subspace in these methods, they may fail to fully capture the intrinsic structure of the data, thus constraining its flexibility in practical applications. In addition to the aforementioned kernel-based methods, numerous works in the field of neural networks focus on classifying missing data. For instance, the responses of neurons in the first hidden layer were replaced with their expected values to minimize the extent of modifications when adapting to various networks~\citep{smieja2018processing}. Later, a neural network architecture called NeuMiss was introduced~\citep{le2020neumiss}. It utilized a Neumann-series approximation of the optimal predictor and demonstrated robustness against various missing data mechanisms. However, as it essentially learns a linear model, its capability may be limited in complex scenarios. It is worth noting that training neural networks typically requires a large amount of data, making it difficult to impute missing values for small-scale data. And in fact, imputing small-scale data is even more critical and meaningful. This is because they inherently provide limited information, and imputation can be used to uncover the underlying structure. On the other hand, it may be challenging to control the outcome of data imputation due to the high degree of freedom during network optimization. 
Furthermore, apart from explorations in algorithmic applications, there are also dedicated efforts focused on theoretical derivations related to missing data. For example, the information-theoretic upper and lower bounds of precision limits for vanilla SVM in the context of learning with missing data were provided~\citep{bullins2016limits}. It was also proven that a predictor designed for complete observations can achieve optimal predictions on incomplete data by utilizing multiple imputation~\citep{josse2019consistency}.

\section{Two-stage Data Imputation}\label{sec3}

\subsection{Preliminaries}\label{sec3.1}

\textbf{Notations}. The set of real numbers is written as $\mathbb{R}$. The set of integers from 1 to $N$ is written as $[N]$. The cardinality of the set $\mathcal{A}$ is written as $|\mathcal{A}|$. We take $a$, $\boldsymbol{a}$, and $\boldsymbol{A}$ to be a scalar, a vector, and a matrix, respectively. Let $\boldsymbol{0}$ and $\boldsymbol{1}$ denote vectors consisting of all zeros and all ones with the appropriate size. The inner product of two vectors in the given space is written as $\langle\cdot, \cdot\rangle$. We take $\mathrm{diag}(\boldsymbol{a})$ to be an operator that extends a vector to a diagonal matrix and $\mathrm{vec}(\mA)$ to be an operator that converts a matrix into a vector by stacking the columns of $\mA$ on top of one another. The Frobenius norm is written as $\|\cdot\|_{\mathrm{F}}$. The set of symmetric matrices is denoted as $\mathcal{S}$, while the set of positive semi-definite (PSD) matrices is denoted as $\mathcal{S}_{+}$. The Hadamard product is written as $\odot$.

\begin{figure}[t]
    \centering
    \includegraphics[width=1\linewidth]{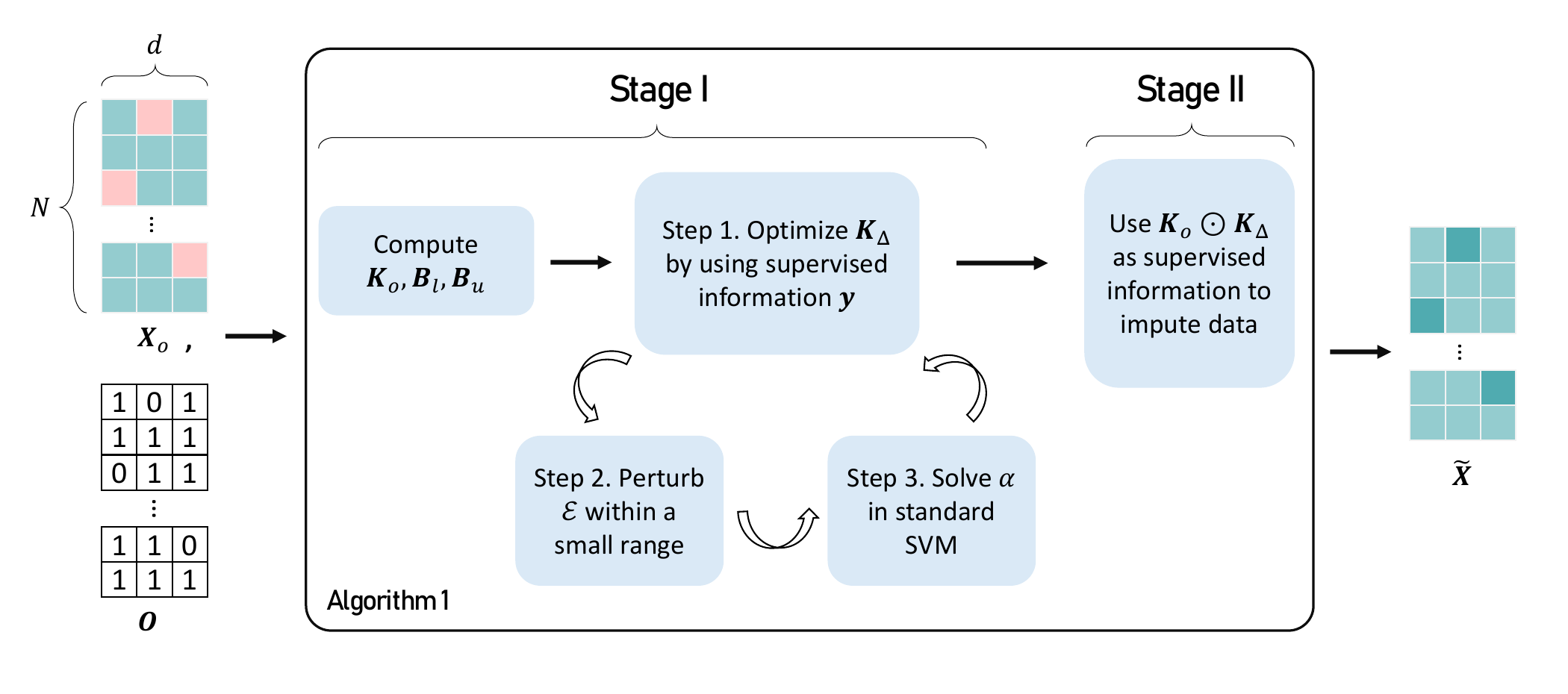}
    \caption{Schematic diagram of the two-stage data imputation framework.}
    \label{Fig: diagram}
\end{figure}

We introduce the vanilla SVM and then demonstrate how to fill in missing data to achieve improved classification performance. In the hard-margin SVM, the objective is to find a hyperplane $\vw^\top \vx+b=0$ that maximally separates the data $\vx\in\mathbb{R}^d$ of different classes with zero training errors. When the data is not linearly separable, the soft-margin SVM is proposed, which allows for some training errors by introducing slack variables denoted as $\xi_i \geq 0$ for each training sample. Meanwhile, to find a more flexible decision boundary, a nonlinear mapping $\phi(\cdot)$ from the $\mathbb{R}^d$ to a reproducing kernel Hilbert space $\mathcal{H}_k$ is introduced, yielding the decision function $f(\vx)=\mathrm{sign}(\vw^\top \phi(\vx)+b)$. With the completed data, the optimization problem can be formulated as 
\begin{equation}
    \begin{aligned}
        \min_{\vw,b,\{\xi_i\}}\quad& \frac{1}{2}\|\vw\|_2^2+C\sum_{i=1}^N\xi_i\\
        \st\quad & y_i(\vw^\top\phi(\vx_i)+b)\geq 1-\xi_i,\ \xi_i \geq 0, \ \forall i\in[N],
    \end{aligned}
    \label{Equ: VanillaSVM}
\end{equation}
where $C\geq 0$ is a hyperparameter that controls the balance between maximizing the margin and minimizing the training errors. For problem (\ref{Equ: VanillaSVM}), previous researchers have proven that we only need to solve its corresponding dual problem:
\begin{equation}
    \begin{aligned}
        \max_{\valpha\in\mathbb{R}^N}\quad& \vone^\top\valpha-\frac{1}{2}\valpha^\top\diag(\vy)\mK\diag(\vy)\valpha\\
        \st\quad & \vy^\top\valpha=0 , \ \vzero\leq \valpha\leq C\vone,
    \end{aligned}
    \label{Equ: DualSVM}
\end{equation}
where $\vy=[y_1;y_2\ ;\ldots; y_N]\in\mathbb{R}^N$ and each entry $K_{i,j}=\langle \phi(\vx_i), \phi(\vx_j) \rangle_{\mathcal{H}_k}$ of the kernel matrix $\mK\in\mathbb{R}^{N\times N}$ represents the similarity between two data. Through a technique known as the kernel trick, this similarity can be computed using a predefined positive definite kernel function $k(\vx_i, \vx_j): \mathbb{R}^d\times \mathbb{R}^d\rightarrow\mathbb{R}$, allowing us to calculate it without knowing the explicit expressions of $\phi$. And the resulting kernel matrix $\mK$ is guaranteed to be PSD. Next, we will introduce our proposed two-stage data imputation method stage by stage. A schematic diagram of the method is shown in Figure~\ref{Fig: diagram}.

\subsection{Stage I: Kernel Matrix Imputation with SVM}\label{sec3.2}

In this stage, our goal is to find a new kernel matrix $\tilde{\mK}$ that further reduces the objective function value in problem~(\ref{Equ: DualSVM}), aiming to improve the classification accuracy of the trained classifier. We propose a novel algorithm that integrates kernel matrix imputation with the classification task. Initially, we compute the observed kernel matrix $\mK_o$, and then optimize an adjustment matrix $\mK_{\Delta}$ with the same dimension to derive $\tilde{\mK}=\mK_o\odot \mK_{\Delta}$ as the optimized kernel matrix. This approach distinguishes itself from previous algorithms by effectively utilizing each observed feature value. Considering the Gaussian kernel function $k_{\gamma}(\vx, \vy) = \exp(-\gamma\|\vx - \vy\|_2^2)$ as an example and recalling the definition of the incomplete dataset $\mathcal{D}=\{\vx_{\vo_i}^i, y_i\}_{i=1}^N$, we can decompose $K_{i,j}$ into an observed part and unknown parts: 
\begin{equation}
    \begin{aligned}
        &K_{i,j}=\mathrm{exp}\left(-\gamma D_{i,j}\right) \ \mathrm{where}\\
        &D_{i,j}=\|\vx_{\vo_i}^i- \vx_{\vo_j}^j\|_2^2\\
        &=\sum_{p\in\vo_i\cap\vo_j}\left(x_p^i-x_p^j\right)^2+\sum_{p\in\vo_i\backslash\vo_j}\left(x_p^i-\ast\right)^2+\sum_{p\in\vo_j\backslash\vo_i}\left(\ast-x_p^j\right)^2+\sum_{p\notin\vo_i\cup\vo_j}\left(\ast-\ast\right)^2.
    \end{aligned}
    \label{Equ: DecoupleKernel}
\end{equation}
Recall that $\ast$ represents an unknown real number. Methods in~\citep{chechik2008max, hazan2015classification} only consider the similarity between the intersecting observed features of two data, i.e., the part related to $p\in\vo_i\cap\vo_j$. In doing so, they disregard the features that are observed in one data but missing in the other, namely $x_p^i$ $(p\in\vo_i\backslash\vo_j)$ and $x_p^j$ $(p\in\vo_j\backslash\vo_i)$. In contrast, our algorithm calculates $\mK_o$ using the first term in the last line of equation (\ref{Equ: DecoupleKernel}) and implicitly incorporates the remaining terms by constraining the potential ranges for $\mK_{\Delta}$. Assuming $\vx$ is min-max normalized within the range of $[0, 1]$, the ranges of the second, third, and fourth terms in equation (\ref{Equ: DecoupleKernel}) are denoted as $[0, \sum_{p\in\vo_i\backslash\vo_j}\max\{(x_p^i)^2, (1-x_p^i)^2\}]$, $[0, \sum_{p\in\vo_j\backslash\vo_i}\max\{(x_p^j)^2, (1-x_p^j)^2\}]$, and $[0, d-|\vo_i\cup\vo_j|]$, respectively. By computing the range of each entry $K_{i,j}=(K_o)_{i,j}\cdot (K_\Delta)_{i,j}$, we can determine the feasible domain of the optimization variable $\mK_{\Delta}$, denoted as $\mB_l\preceq \mK_{\Delta} \preceq \mB_u$. In Stage II and the experimental sections, we will continue to use the Gaussian kernel as an example. However, it is noted that our algorithm can be applied to various kernel functions, as long as their computations can be decoupled for each feature. This decoupling can be expressed as
\begin{equation}
    k(\vx,\vy) = f\left(\sum_{p=1}^dg\left(x_i,y_i\right)\right),
\end{equation}
where the range of the binary function $g(\cdot, \cdot)$ needs to be bounded when both inputs in the range $[0, 1]$, and $f(\cdot)$ needs to be a monotonic function on the range of $\sum_{p=1}^dg\left(x_i,y_i\right)$. Many commonly used kernels exhibit this property, including the polynomial kernel, sigmoid kernel, $\chi^2$ kernel, and more. For a detailed derivation of the decoupling for these kernels, please refer to the~\ref{secA}.

By leveraging the supervision information and performing alternating optimization between the kernel matrix and the classifier, we will obtain a kernel matrix that yields better classification performance. Meanwhile, due to the flexibility of this approach, we aim to ensure the robustness of the classifier concurrently, thus mitigating the risk of overfitting. Therefore, based on the concept of robust optimization~\citep{xu2009robustness,cooper2022believe}, we introduce a perturbation variable denoted as $\mathcal{E}\in\mathbb{R}^{N \times N}$ to guarantee that the classifier performs well not only on $\tilde{\mK}$ but also on all possible outcomes within a norm sphere surrounding $\tilde{\mK}$. Finally, we formulate the following optimization problem for Stage I: 
\begin{equation}
    \begin{aligned}
        \min_{\mK_{\Delta}}\max_{\valpha, \mathcal{E}}\quad& \mathcal{L}=\vone^\top\valpha-\frac{1}{2}\valpha^\top\diag(\vy)\left(\mK_o\odot\mK_{\Delta}\odot\mathcal{E}\right)\diag(\vy)\valpha+\eta\|\mK_{\Delta}-\vone\vone^\top\|_{\mathrm{F}}^2\\
        \st\quad & \mB_l\preceq \mK_{\Delta} \preceq \mB_u,\  \mK_{\Delta}\in\mathcal{S}_{+},\\
        & \|\mathcal{E}-\vone\vone^\top\|_{\mathrm{F}}^2 \leq r^2,\ \mathcal{E}\in\mathcal{S}_{+},\\
        & \vy^\top\valpha=0 , \ \vzero\leq \valpha\leq C\vone,
    \end{aligned}
    \label{Equ: Stage1}
\end{equation}
where the regularization parameters $\eta$ and $r$ are introduced to control the range of modification for the observed kernel matrix $\mK_o$ and the intensity of the perturbation, respectively. By imposing PSD constraints on the variables $\mK_{\Delta}$ and $\mathcal{E}$, it is ensured that the optimized kernel matrix remains PSD. As the value of $\eta$ increases, $\mK_{\Delta}$ tends to approach the all-one matrix. In this scenario, the model approximates a state where the missing values are filled with zeros. To solve the optimization problem outlined in (\ref{Equ: Stage1}), we propose an alternating optimization algorithm consisting of three steps.

\textbf{Step 1.} Optimizing $\mK_{\Delta}$ with fixed $\mathcal{E}$ and $\valpha$. In this step, problem (\ref{Equ: Stage1}) with respect to $\mK_{\Delta}$ is equivalent to 
\begin{equation}
    \begin{aligned}
        \min_{\mK_{\Delta}}\quad& -\frac{1}{2}\valpha^\top\diag(\vy)\left(\mK_o\odot\mK_{\Delta}\odot\mathcal{E}\right)\diag(\vy)\valpha+\eta\|\mK_{\Delta}-\vone\vone^\top\|_{\mathrm{F}}^2\\
        \st\quad & \mB_l\preceq \mK_{\Delta} \preceq \mB_u,\  \mK_{\Delta}\in\mathcal{S}_{+}.
    \end{aligned}
\label{Equ: SVM_S1}
\end{equation}
The problem mentioned above is a semi-definite program, with inequality constraints imposed on each element of the matrix $\mK_{\Delta}$, making it generally computationally expensive to solve. To alleviate the computational burden, we instead propose a strategy for finding an approximate solution to the optimal solution of the problem (\ref{Equ: SVM_S1}). Firstly, we disregard the inequality constraints and the PSD constraint and solve the problem using the following formulation: 
\begin{equation}\label{equ: 7}
    \begin{aligned}
        \hat{\mK}_{\Delta}&\coloneqq\mathop{\arg\min}_{\mK_{\Delta}}\left\{-\frac{1}{2}\valpha^\top\diag(\vy)\left(\mK_o\odot\mK_{\Delta}\odot\mathcal{E}\right)\diag(\vy)\valpha+\eta\|\mK_{\Delta}-\vone\vone^\top\|_{\mathrm{F}}^2\right\}\\
        &= \vone\vone^\top+\frac{1}{4\eta}\diag(\valpha\odot\vy) \left(\mK_o\odot \mathcal{E}\right)\diag(\valpha\odot\vy).
    \end{aligned}
\end{equation}
Since $\hat{\mK}_{\Delta}$ is a real symmetric matrix, we can perform eigenvalue decomposition on the matrix as $\hat{\mK}_{\Delta} \coloneqq\mU\mSigma\mU^\top$. Subsequently, within the feasible domain defined by $\mB_l\preceq \mK_{\Delta} \preceq \mB_u,\  \mK_{\Delta}\in\mathcal{S}_{+}$, we adjust the eigenvalues of the matrix to find a solution that is closest to $\hat{\mK}_{\Delta}$, i.e., 
\begin{equation}\label{Equ: UtoK}
    \begin{aligned}
        \min_{\vd\in\mathbb{R}^N}\quad& \left\|\mU\diag(\vd)\mU^\top-\hat{\mK}_{\Delta}\right\|_{\mathrm{F}}^2\\
        \st\quad & \mB_l\preceq \mU\diag(\vd)\mU^\top \preceq \mB_u,\  \vd\succeq \vzero.
    \end{aligned}
\end{equation}
We define $\mU=\left[\vu_1\ \vu_2\ \ldots\ \vu_N\right]$ and the $(i,j)$-th entry of $\vu_k\vu_k^\top\in\mathbb{R}^{N\times N}$ as $\left(\vu_k\vu_k^\top\right)_{i,j}\coloneqq u_{i,j}^k\ (k\in[N])$. By doing so, we can reformulate the problem (\ref{Equ: UtoK}) as below,
\begin{equation}
    \begin{aligned}
        \min_{\vd\in\mathbb{R}^N}\quad& \left\|\mV^\top\vd-\mathrm{vec}(\hat{\mK}_{\Delta})\right\|_{2}^2\\
        \st\quad & \mathrm{vec}(\mB_l)\preceq \mV^\top\vd \preceq \mathrm{vec}(\mB_u),\  \vd\succeq \vzero,
    \end{aligned}
\end{equation}
where $\mV= \left[\vv_{1,1}\ \vv_{2,1}\ \ldots\ \vv_{N,1}\ \vv_{1,2}\ \vv_{2,2}\ \ldots\ \vv_{N,N}\right]\in\mathbb{R}^{N\times N^2}$ and for each $\vv_{i,j} = \left[u_{i,j}^1;\ u_{i,j}^2 ;\ \ldots\ ;\  u_{i,j}^N\right]\in\mathbb{R}^N$. To further simplify the optimization problem mentioned above, we introduce an auxiliary matrix $\mG \coloneqq \left[\mV;\ \mV;\ \textnormal{-}\mI\right]\in\mathbb{R}^{N\times (2N^2+N)}$ and an auxiliary vector $\vh\coloneqq \left[\mathrm{vec}(\mB_u);\ \textnormal{-}\mathrm{vec}(\mB_l);\ \vzero\right]\in\mathbb{R}^{2N^2+N}$, resulting in the following problem:
\begin{equation}\label{Equ: find_d}
    \begin{aligned}
        \min_{\vd\in\mathbb{R}^N}\quad& \vd^\top \mV\mV^\top\vd - 2\left[\mV\mathrm{vec}(\hat{\mK}_{\Delta})\right]^\top\vd\\
        \st\quad &\mG^\top\vd \preceq \vh.
    \end{aligned}
\end{equation}
Therefore, we need to solve a convex linear constrained quadratic programming problem, which can be efficiently accomplished using existing convex optimization toolboxes, such as the $\texttt{quadprog}$ function in MATLAB. After obtaining the optimal solution $\vd^\ast$, we can then derive the final approximate solution for the problem (\ref{Equ: SVM_S1}):
\begin{equation}
    \mK_{\Delta}^\ast=\mU\diag(\vd^\ast)\mU^\top.
    \label{Equ: OptKD}
\end{equation}

\textbf{Step 2.} Optimizing $\mathcal{E}$ with fixed $\mK_{\Delta}$ and $\valpha$. In this step, we aim to perturb the optimization variables within a small range and maximize the objective function value in the problem (\ref{Equ: Stage1}). For this problem, similarly, to avoid directly solving the semi-definite program, we disregard the PSD constraint and propose an approximate optimization approach for $\mathcal{E}$ as follows,
\begin{equation}
    \begin{aligned}
        \min_{\mathcal{E}}\quad& \frac{1}{2}\valpha^\top\diag(\vy)\left(\mK_o\odot\mK_{\Delta}\odot\mathcal{E}\right)\diag(\vy)\valpha\\
        \st\quad & \|\mathcal{E}-\vone\vone^\top\|_{\mathrm{F}}^2 \leq r^2.
    \end{aligned}
    \label{Equ: SVM_S2}
\end{equation}
Due to the linearity of the objective function with respect to $\mathcal{E}$ and the fact that the constraint defines a compact convex set, it can be inferred that the optimal solution will lie on the boundary of the constraint. Accordingly, we can relax the constrained optimization problem stated in (\ref{Equ: SVM_S2}) to an unconstrained problem:
\begin{equation}
    \min_{\mathcal{E}}\quad \frac{1}{2}\valpha^\top\diag(\vy)\left(\mK_o\odot\mK_{\Delta}\odot\mathcal{E}\right)\diag(\vy)\valpha+\rho\|\mathcal{E}-\vone\vone^\top\|_{\mathrm{F}}^2.
    \label{Equ: SVM_S2_nost}
\end{equation}
After deriving the closed-form solution for this problem, we utilize a non-expansive operator $\mathcal{P}_{+}(\cdot):\mathcal{S}^N\rightarrow \mathcal{S}_{+}^N$ to project the solution onto the space of PSD matrices~\citep{cai2010singular,xu2023fast}. The operator $\mathcal{P}_{+}(\mX)$ performs eigenvalue decomposition on $\mX=\mV\mSigma\mV^\top$ and sets any negative eigenvalues to 0, resulting in the new matrix $\hat{\mX}=\mV\mSigma_{+}\mV^\top$. By defining $\Gamma\coloneqq \diag(\valpha\odot\vy) \left(\mK_o\odot \mK_\Delta\right)\diag(\valpha\odot\vy)$, we obtain the optimal PSD approximation solution for the problem (\ref{Equ: SVM_S2}):
\begin{equation}
    \mathcal{E}^\ast=\mathcal{P}_{+}\left(\vone\vone^\top-\frac{1}{4\rho}\Gamma\right).
    \label{Equ: OptE}
\end{equation}

\textbf{Step 3.} Optimizing $\valpha$ with fixed $\mK_{\Delta}$ and $\mathcal{E}$. Given $\mK_{\Delta}$ and $\mathcal{E}$, the optimization problem is reduced to the standard SVM problem, which can be solved using various methods. In this case, we employ the gradient descent method to update $\valpha$ in each iteration and utilize the Adam optimizer~\citep{kingma2014adam} to adjust the learning rate dynamically. After obtaining the updated variable $\hat{\valpha}$, we proceed to project it onto the feasible set. This is done by first clipping it to the range $[0,C]$ and then calculating $\valpha^\ast=\hat{\valpha}-\frac{\vy^\top\hat{\valpha}}{N}\vy$ as the final solution for this step.

Additionally, for the non-trivial min-max problem with box and PSD constraints presented in problem~(\ref{Equ: Stage1}), although it can be solved through the above three alternating optimization steps, its convergence remains unclear. In \ref{secB}, we provide a comprehensive review of the entire procedure, including detailed definitions of all notations, and then, based on several lemmas, present the following theorem concerning the norm difference of the $\mK_\Delta$ parameters during iterations. 
\begin{theorem} 
    When $t$ denotes the iteration index, the difference in $\mK_\Delta$ between two consecutive updates can be bounded by
    \begin{equation}
        \left\|\mK_\Delta^{(t+1)} - \mK_\Delta^{(t)} \right\|_\mathrm{F} \leq L \left\|\mK_\Delta^{(t)}-\mK_\Delta^{(t-1)}\right\|_\mathrm{F} + R^{(t)},
    \end{equation}
    where $L = \frac{1}{4\eta} N^2C^4\kappa^2$ and $R^{(t)} = \frac{1}{2\eta}\nu^{(t)}N^2C^3\kappa^2 b_{\mathrm{max}}\left(1+C\kappa b_{\mathrm{max}} + \frac{NC^3\kappa\lambda b_{\mathrm{max}}^2}{4\rho}\right)+ \frac{1}{2\eta} \nu^{(t)}NC\kappa$ \\$\left(1+\frac{NC^2\kappa b_{\mathrm{max}}}{4\rho}\right)\left(1+C\kappa b_{\mathrm{max}} +\frac{NC^3\kappa\lambda b_{\mathrm{max}}^2}{4\rho}\right)$.
\end{theorem}
\begin{remark}
    In our algorithm, $\eta$, $\rho$, and $\nu^{(t)}$ control the optimization step sizes for variables $\mK_\Delta$, $\mathcal{E}$, and $\valpha$ in each iteration, respectively. When we set $\eta >\frac{N^2C^4\kappa^2}{4}$, $\rho =\mathcal{O}\left(NC^3\kappa\lambda b_{\mathrm{max}}^2\right)$, and ensure that $\nu^{(t)}=\mathcal{O}(1/t)$ holds, it will follows that $L<1$ and $R^{(t)}\rightarrow 0$ as the iteration index $t$ increases. Since the actual upper bound may be smaller, we thus establish a sufficient condition that guarantees the convergence of the aforementioned alternating optimization.
\end{remark}

\subsection{Stage II: Data Imputation with the Given Matrix}\label{subsec3.3}

In the previous stage, we obtained the optimized kernel matrix $\tilde{\mK}=\mK_o\odot \mK_{\Delta}$ by incorporating supervised information of the data during SVM training. In this stage, we will utilize the matrix $\tilde{\mK}$ to perform data imputation. We redefine the incomplete dataset $\mathcal{D}=\{\vx_{\vo_i}^i, y_i\}_{i=1}^N$ as $\mathcal{D}=\{\mX_o,\mY,\mO\}$, where $\mX_o \coloneqq [\vx_{\vo_1}^1\; \vx_{\vo_2}^2 \; \ldots \; \vx_{\vo_N}^N]\in\mathbb{R}^{d\times N}$, $\mY\in\mathbb{R}^N$ and $\mO\in\{0,1\}^{d\times N}$ 
is used to indicate which features are missing (represented by 0) and which features are observed (represented by 1).
Next, we will compute $\Delta\mX\coloneqq [\Delta\vx_1\; \Delta\vx_2 \; \ldots \; \Delta\vx_N]\in\mathbb{R}^{d\times N}$ by using the entries of the trained kernel matrix $\tilde{K}_{i,j}$ as supervisory information. The goal is to minimize the discrepancy between the optimized kernel matrix calculated from the imputed data $\tilde{\mX}=\mX_o+\Delta\mX$ and the original kernel matrix $\tilde{\mK}$ through regression, i.e,
\begin{equation}
    \min_{\{\Delta \vx_i\}_{i=1}^{N}}\quad \sum_{i}\sum_{j}\left[\mathrm{exp}\left(-\gamma\left\|\vx_{\vo_i}^i - \vx_{\vo_j}^j + \Delta\vx_i -\Delta\vx_j\right\|_2^2\right)-\tilde{K}_{i,j}\right]^2.
\end{equation}
By replacing the objective function and imposing location and range constraints on the imputation results, we obtain the following optimization problem for Stage II:
\begin{equation}
    \begin{aligned}
        \min_{\Delta\mX}\quad& \sum_{i}\sum_{j}\left[\left\|\vx_{\vo_i}^i - \vx_{\vo_j}^j + \Delta\vx_i -\Delta\vx_j\right\|_2^2+\frac{1}{\gamma}\mathrm{ln}\left({\tilde{K}}_{i,j}\right)\right]^2\\
        \mathrm{s.t.}\quad& \Delta\mX \odot \mO = \vzero,\\
        &\vzero \preceq \mX_o+\Delta\mX \preceq \vone.
    \end{aligned}
    \label{Equ: Stage2}
\end{equation}

The essence of this task involves solving a nonlinear system of equations. By defining the missing ratio of data features as $m$, the aforementioned system consists of $N(N-1)/2$ equations and $Ndm$ variables. Although the objective function in problem (\ref{Equ: Stage2}) is non-convex to $\Delta\mX$, the entries $\{K_{i,j}\}_{i,j=1}^N$ still provide abundant supervision information for the data imputation process in scenarios where the number of observed features $(d)$ is significantly smaller than the total number of data $(N)$ and the missing ratio $(m)$ ranges between 0 and 1. To address this, we employ the BCD method to solve the optimization problem mentioned above. In each iteration, we select a specific column from $\Delta\mX$ as a variable while keeping the remaining columns constant. We update the variable iteratively until the maximum number of iterations is reached or the change in the objective function is smaller than the predefined threshold. The accuracy of this solution is further validated by the experimental results presented later in this paper. The complete two-stage data imputation framework is summarized in Algorithm~\ref{Algo}.

\hypertarget{time}{\textcolor{red}{In addition, we are equally concerned about the time complexity of the algorithm. First, we analyze Stage I. When computing the preliminary results of $\hat{\mK}_\Delta$, $\hat{\mathcal{E}}$, and $\hat{\valpha}$ (excluding the projection operation), the relevant Hadamard products and matrix-vector multiplications require a complexity of $\mathcal{O}(N^2)$. As for $\mK_\Delta$ and $\mathcal{E}$, their subsequent PSD projection operators and quadratic programming demand a higher complexity of $\mathcal{O}(N^3)$. Next, for Stage II, the complexity required for single-column updates $\Delta\vx_i$ is $\mathcal{O}(Nd)$, so traversing the complete $\Delta\mX$ once would require $\mathcal{O}(N^2d)$. Therefore, in the scenario we are concerned with (where $N\gg d$), the computational cost of Algorithm~\ref{Algo} in each iteration will be dominated by $\mathcal{O}(N^3)$, where the empirical number of iterations can be referenced from the results in Figure~\ref{Fig: converge}.}}

\begin{algorithm}[tbp]
    \caption{Two-Stage Data Imputation Based on Support Vector Machine}
    \label{Algo}
    \begin{algorithmic}[1]
        \Require the incomplete dataset $\mathcal{D}=\{\mX_o,\mY,\mO\}$, the parameters $C$, $\gamma$, $\eta$, and $\rho$.
        \Ensure the imputed data $\tilde{\mX}$ and the combination coefficient $\valpha$.
        \State \textbf{Stage I:}
        \State Compute $\mK_o$, $\mB_l$, and $\mB_u$. Initialize $\mK_{\Delta}^{(0)}=\mathcal{E}^{(0)}=\vone\vone^\top$, $\valpha^{(0)}=\frac{C}{2}\vone$, and $t_1=t_2=0$.
        \Repeat 
        \State Solve problem (\ref{Equ: find_d}) to obtain $d^\ast$ and update $\mK_{\Delta}^{(t_1)}$ by equation (\ref{Equ: OptKD}).
        \State Update $\mathcal{E}^{(t_1)}$ by equation (\ref{Equ: OptE}).
        \State Update $\valpha$ using gradient descent and project it onto the feasible set.
        \State $t_1=t_1+1$.
        \Until the stop criterion is satisfied.
        \State Compute $\tilde{\mK}=\mK_o\odot\mK_{\Delta}$.
        \State \textbf{Stage II:}
        \State Initialize $\Delta\vx_i^{(0)}=\frac{1}{2}\vone$.
        \Repeat
        \For{$i=1$ to $N$}
        \State Fix $\{\Delta\vx_j^{(t_2-1)}: j\neq i\}$ and update $\Delta\vx_i^{(t_2)}$ in problem (\ref{Equ: Stage2}).
        \EndFor
        \State $t_2 = t_2 + 1$.
        \Until the stop criterion is satisfied.
    \end{algorithmic}
\end{algorithm}

\section{Numerical Experiments}
\subsection{Experimental Settings}
\textbf{Datasets and preprocessing.} We chose seven real-world datasets from libsvm~\citep{CC01a} and UCI~\citep{UCIdata}, namely \textit{australian}, \textit{german}, \textit{heart}, \textit{pima (a.k.a. diabetes)}, \textit{wine}, \textit{cylinder}, and \textit{horse}. The details of these datasets are shown in Table~\ref{Tab: data}. The datasets were then divided into three subsets: a training set, a complete holdout set for parameter selection, and a complete test set for evaluating and comparing the final results of the algorithms. The split was performed in a 4:3:3 ratio. Additionally, for the training set of the first five complete datasets, we constructed the missing data by randomly removing a given percentage of the features to evaluate the performance of various algorithms under different levels of missingness. The missing ratio of a dataset was defined as
\begin{equation*}
    m\coloneqq\frac{\text{\# The missing features}}{\text{\# The total features}}.
\end{equation*}
For the last two datasets that have missing values, we directly utilized their inherent missing patterns to test the performance of algorithms in the absence of prior knowledge about the distribution of missing values.
For preprocessing, we scaled $\vx_i$ to $[0, 1]$ using the observed features and scaled $y_i$ to $\{-1, 1\}$.

\begin{table}[tb]
    \centering
    \caption{Details of the seven libsvm and UCI datasets.}
    \label{Tab: data}
    \setlength{\tabcolsep}{2.5mm}
    \begin{tabular}{ccc}
    \toprule[1.5pt]
    datasets & \# Features ($d$) & \# Instances ($N$) \\ \midrule[1pt]
    Australian & 14 & 690  \\
    German & 21 & 1000  \\
    Heart & 13 & 270  \\
    Pima & 8 & 768  \\ 
    Wine & 11 & 4898  \\
    \midrule[0.5pt]
    Cylinder & 35 & 512 \\
    Horse & 25 & 368  \\
    \bottomrule[1.5pt]
    \end{tabular}
\end{table}

\textbf{Compared methods and parameters settings.} We selected a basic method along with five advanced methods for comparison with our proposed framework. To quantify the performance of the individual methods, we compared their classification accuracy on the test dataset. All experiments were repeated 10 times, and the average accuracy of each method was reported. The implementation was carried out using MATLAB on a machine with an Intel$^\circledR$ Core$^\text{TM}$ i7-11700KF CPU (3.60 GHz) and 32GB RAM. The source code is available in \url{https://github.com/Yruikk/TwoStageDataImputation}.
\begin{itemize}
    \item \textbf{MI}: The missing values are imputed by replacing them with the mean value of the corresponding features in the training set. We chose the penalty parameter $C \in \{2^{-5},\ 2^{-4},\ \ldots\ ,2^{5}\}$ and the bandwidth in the Gaussian kernel $\gamma\in \{2^{-5},\ 2^{-4},\ \ldots\ ,2^{5}\}$ using the holdout set.
    \item \textbf{GEOM}~\citep{chechik2008max}: An iterative framework for optimizing a non-convex problem was introduced in this work. The main objective is to maximize the margin within the relevant subspace of the observed data. The parameters $C$ and $\gamma$ are chosen the same as in MI, and the iteration round $t$ is chosen from the set $\{2,3,4,5\}$.
    \item \textbf{KARMA}~\citep{hazan2015classification}: This method proposed a new kernel function for the missing data $k_{\beta}(\cdot,\cdot):(\mathbb{R}\cup \{\ast\})^d\times (\mathbb{R}\cup\{\ast\})^d\rightarrow\mathbb{R}$.  We chose $C \in \{2^{-5},\ 2^{-4},\ \ldots\ ,2^{5}\}$ and $\beta\in\{1,2,3,4\}$ using the holdout set.
    \item \textbf{genRBF}~\citep{smieja2019generalized}: It derived an analytical formula for the expected value of the radial basis function kernel over all possible imputations. The parameters $C$ and $\gamma$ were chosen the same as in MI.
    \item \textbf{NewImp}~\citep{chen2024rethinking}: This method employs the Wasserstein gradient flow framework and performs data imputation based on a diffusion model. For the hyper-parameters $h$, $\mathrm{HU}_{\mathrm{score}}$, $\lambda$, $\eta$, we follow the grid range described in their paper. Additionally, for SVM, $C$ and $\gamma$ were chosen the same as in MI.
    \item \textbf{Ours}: To avoid excessive parameter tuning, we utilized $C_{\mathrm{MI}}$ and $\gamma_{\mathrm{MI}}$ from MI and further selected $C\in\{C_{\mathrm{MI}}\cdot2^i,\ i=-1,0,1\}$ and $\gamma\in\{\gamma_{\mathrm{MI}}\cdot2^i,\ i=-1,0,1\}$. Additionally, we empirically set $\eta=\|\valpha_{\mathrm{MI}}\|_2$ and $\rho= 5C/m$ during each iteration.
\end{itemize}

\subsection{Experimental Results}
\subsubsection{Results of Data Imputation from Given Kernel Matrix}\label{sec: 4.2.1}

\begin{table}[tb]
    \centering
    \caption{Results of data imputation performance using a given kernel matrix with different Gaussian kernel bandwidths and missing ratios.}
    \label{Tab: KtoX}
    \centering
    \begin{tabular}{cccccc}
    \toprule[1.5pt]
    Parameter & $m$ & $(e_{\mX})_{\text{max}}$ & $(e_{\mX})_{\text{mean}}$ & $(e_{\mK})_{\text{max}}$ & \multicolumn{1}{c}{$(e_{\mK})_{\text{mean}}$} \\ \midrule[1pt]
    \multirow{2}{*}{$\gamma=1$} & $10\%$ & $3.46\mathrm{e}{-4}$ & $3.00\mathrm{e}{-7}$ & $5.95\mathrm{e}{-5}$ & $8.85\mathrm{e}{-8}$ \\
     & $90\%$ & $1.27\mathrm{e}{-1}$ & $6.71\mathrm{e}{-4}$ & $5.57\mathrm{e}{-2}$ & $8.28\mathrm{e}{-5}$ \\ \midrule[1pt]
    \multirow{2}{*}{$\gamma=\frac{1}{32}$} & $10\%$ & $2.75\mathrm{e}{-4}$ & $2.88\mathrm{e}{-7}$ & $1.32\mathrm{e}{-5}$ & $4.52\mathrm{e}{-8}$ \\
     & $90\%$ & $1.52\mathrm{e}{-1}$ & $1.56\mathrm{e}{-3}$ & $3.70\mathrm{e}{-3}$ & $5.83\mathrm{e}{-5}$ \\ \midrule[1pt]
    \multirow{2}{*}{$\gamma=32$} & $10\%$ & $7.30\mathrm{e}{-4}$ & $4.74\mathrm{e}{-7}$ & $3.08\mathrm{e}{-5}$ & $6.81\mathrm{e}{-9}$ \\
     & $90\%$ & $3.81\mathrm{e}{-1}$ & $3.75\mathrm{e}{-3}$ & $1.03\mathrm{e}{-1}$ & $2.44\mathrm{e}{-5}$ \\ 
    \bottomrule[1.5pt]
    \end{tabular}
\end{table}

In order to ensure the operation of the entire framework, we first evaluated the performance of Stage II. We conducted experiments on the \textit{heart} dataset, considering various bandwidths of the Gaussian kernel and different missing ratios. We computed the complete kernel matrix $\mK_{\mathrm{gt}}$ based on the complete data $\mX_{\mathrm{gt}}$ and the chosen bandwidth $\gamma$. Subsequently, we randomly removed $Ndm$ features from the complete data, resulting in $\mX_{\mathrm{miss}}$. Next, utilizing the algorithm from Stage II in our framework, we performed data imputation on $\mX_{\mathrm{miss}}$ using the complete kernel matrix $\mK_{\mathrm{gt}}$, which yielded $\tilde{\mX}$. 
In addition to caring about the quality of data imputation, we also focus on the quality of the kernel matrix used in predicting the class of new data. Therefore, we defined four evaluation metrics to assess the error between $\tilde{\mX}$ and $\mX_{\mathrm{gt}}$, as well as the error between $\tilde{\mK}$ and $\mK_{\mathrm{gt}}$: $(e_{\mX})_{\text{max}}$, $(e_{\mX})_{\text{mean}}$, $(e_{\mK})_{\text{max}}$, and $(e_{\mK})_{\text{mean}}$. We then conducted tests on our data imputation algorithm using different bandwidths of the Gaussian kernel and missing ratios. The results of these tests were reported in Table~\ref{Tab: KtoX}. We observed that at a low missing ratio $(m=10\%)$, both the imputed data and kernel matrix closely matched the ground truth. Even at a high missing ratio $(m=90\%)$, although there were some imputed features that deviated significantly from the true values, the overall performance of the algorithm in terms of average imputation remained highly accurate. 
Additionally, we found that when the Gaussian kernel parameter was particularly small $(\gamma=1/32)$ or large $(\gamma=32)$, the true kernel matrix tended to be a matrix of all ones or an identity matrix. In these scenarios, although there may be larger errors in the imputed data compared to when $\gamma=1$, the kernel matrix used for actual predictions still maintained a good level of accuracy.

\subsubsection{Classification Results on Artificial datasets}
\label{sec: 4.2.2}
\begin{figure}[t]
    \centering
    \begin{subfigure}{0.48\textwidth}
        \includegraphics[width=\textwidth]{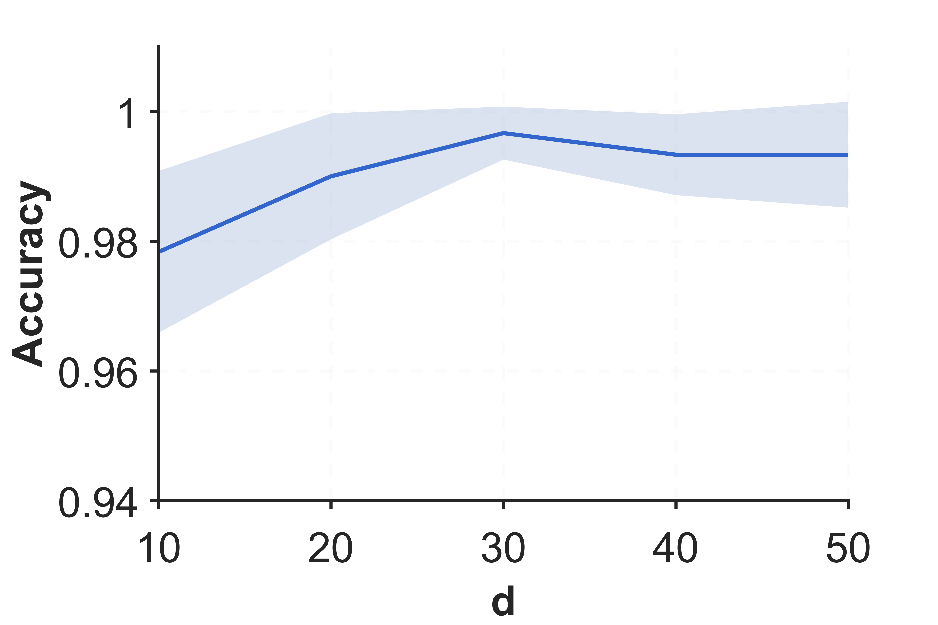}
    \end{subfigure}
    \hfill
    \begin{subfigure}{0.48\textwidth}
        \includegraphics[width=\textwidth]{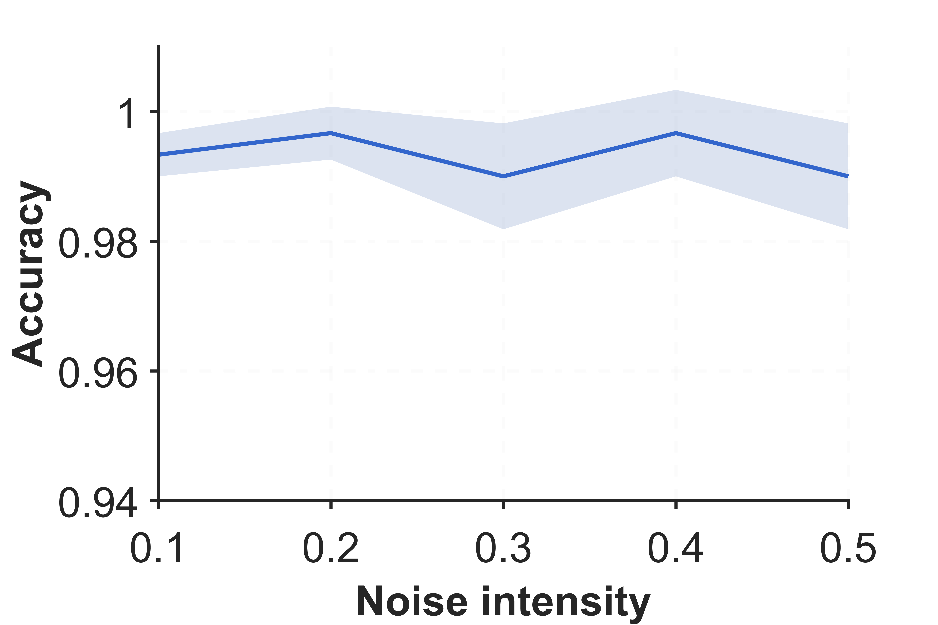}
    \end{subfigure}
    
    \vspace{1em} 

    \begin{subfigure}{0.48\textwidth}
        \includegraphics[width=\textwidth]{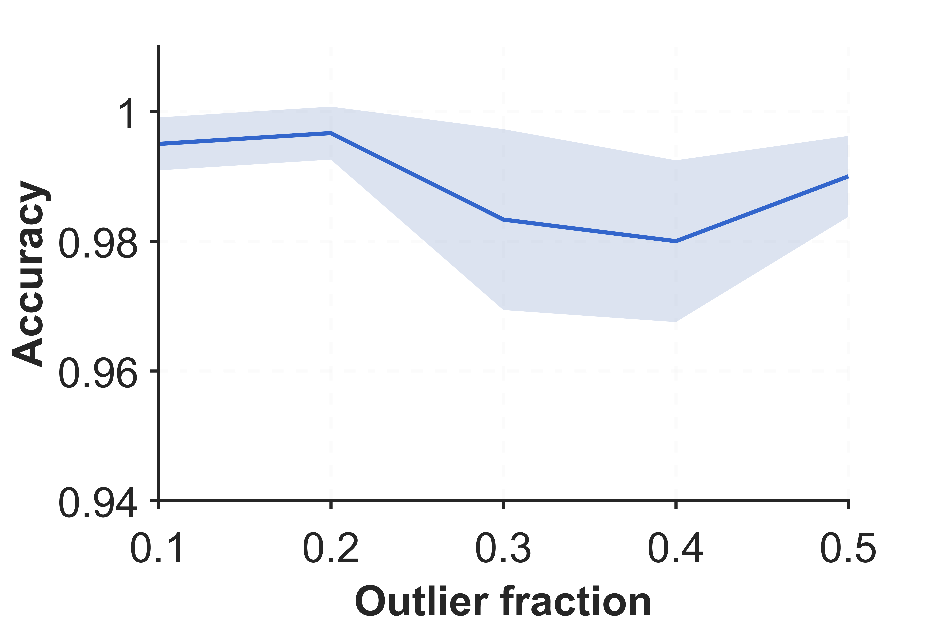}
    \end{subfigure}
    \hfill
    \begin{subfigure}{0.48\textwidth}
        \includegraphics[width=\textwidth]{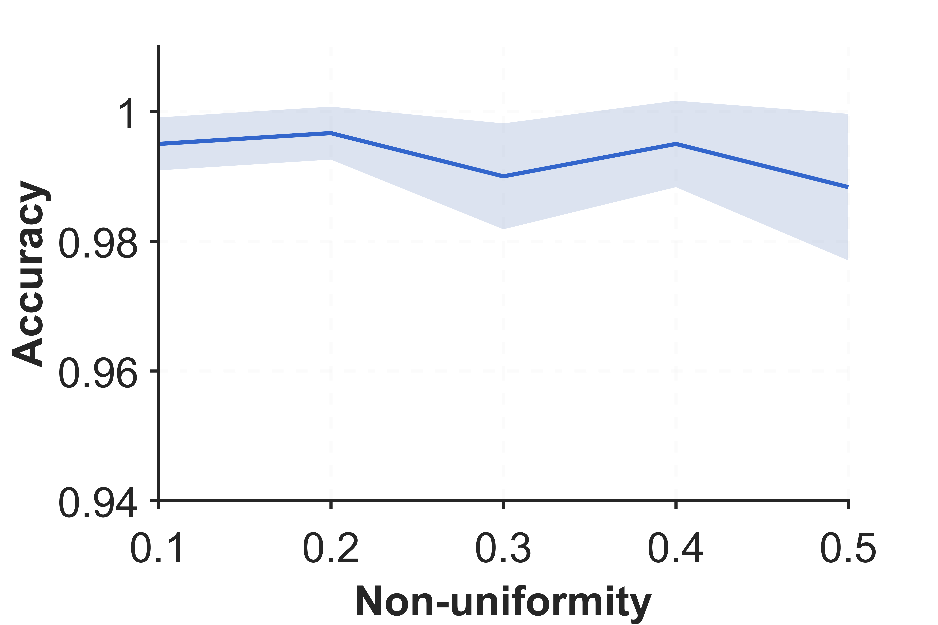}
    \end{subfigure}

    \caption{The performance of the proposed method (mean $\pm$ std) on artificial datasets with varying dimensions ($d$), noise intensity, outlier fraction, and non-uniformity.}
    \label{fig:synthetic}
\end{figure}

To validate the robustness of our method first, we conducted evaluations using artificially generated datasets with controlled characteristics. These datasets were designed to systematically investigate the influence of four key factors: dimensionality, noise intensity, extreme outliers, and non-uniform distributions. Specifically, we generated two classes of data as follows: $\mX_1 \sim \mathcal{N}(\vmu_1, \mSigma_1)$ and $\mX_2 \sim \mathcal{N}(\vmu_2, \mSigma_2)$, where $\vmu_1 = -\vone_d$, $\vmu_2 = \vone_d$  and $\mSigma_1 = \mSigma_2 = \diag(1 + \mathrm{non\_uniformity} \cdot \vu_d)$, with $\vu_d \sim \mathrm{Uniform}(\vzero_d,\vone_d)$. To simulate noise intensity, Gaussian noise was added to each feature. The noisy dataset is represented as $\mX_{\mathrm{noisy}} = \mX + \mathrm{noise\_intensity} \cdot \mathcal{N}(0, 1)$, where $\mX$ represents the combined dataset from both classes. The effect of extreme outliers was simulated by replacing a fraction of the samples with outlier values, generated as $\mX_{\mathrm{outliers}} = \mX + 10 \cdot (\mU - 0.5)$, where $\mU_{i,j}$ is drawn from a uniform distribution $\mathrm{Uniform}(0, 1)$. 
The effects of these parameters on imputation and subsequent classification tasks are summarized in Figure~\ref{fig:synthetic}. For all experiments, the default parameter values were set as $N=400$, \(d = 30\), \(\mathrm{noise\_intensity} = 0.2\), \(\mathrm{outlier\_fraction} = 0.1\), \(\mathrm{non\_uniformity} = 0.5\), and \(C = \gamma = 1\) for the SVM classifier. Our results show that the proposed method demonstrates strong robustness across different scenarios. As expected, increasing the noise intensity, fraction of outliers, and non-uniformity slightly reduces the accuracy and increases the variance. However, the overall mean accuracy remains relatively stable. 

\begin{table*}[tb]
\caption{Comparison of classification accuracy (mean $\pm$ std) with a baseline and three state-of-the-art methods on four real datasets. We conducted testing on the scenarios with different missing ratios of the data. The best performance is highlighted in bold and the ``$\bullet$'' indicates its statistical significance compared to other methods via paired t-test at the 5\% significance level.}
\label{Tab: real-world results}
\centering
\resizebox{\linewidth}{!}{
\begin{tabular}{cccccc}
\toprule[1.5pt]
\multirow{2.5}{*}{\makecell[c]{datasets}} & \multirow{2.5}{*}{\makecell[c]{Methods}} & \multicolumn{4}{c}{$m$} \\ \cmidrule[0.75pt]{3-6} 
 &  & $20\%$ & $40\%$ & $60\%$ & $80\%$ \\ \midrule[1pt]
\multirow{6}{*}{\makecell[c]{Australian}} & MI & 0.864 $\pm$ 0.021 & 0.851 $\pm$ 0.028 & 0.807 $\pm$ 0.047 & 0.678 $\pm$ 0.049 \\
 & GEOM & 0.849 $\pm$ 0.025 & 0.831 $\pm$ 0.037 & 0.754 $\pm$ 0.050 & 0.632 $\pm$ 0.043 \\
 & KARMA & 0.861 $\pm$ 0.020 & 0.848 $\pm$ 0.022 & 0.831 $\pm$ 0.045& 0.712 $\pm$ 0.046\\
 & genRBF & \textbf{0.867} $\pm$ \textbf{0.016} & 0.864 $\pm$ 0.023 & 0.796 $\pm$ 0.041 & 0.704 $\pm$ 0.058\\
 & NewImp & 0.865 $\pm$ 0.020 & 0.852 $\pm$ 0.031 & 0.816 $\pm$ 0.026 & 0.777 $\pm$ 0.057\\
 & Ours & 0.867 $\pm$ 0.023 & \textbf{0.867} $\pm$ \textbf{0.021} & \textbf{0.849} $\pm$ \textbf{0.024} $\bullet$ & \textbf{0.857} $\pm$ \textbf{0.025} $\bullet$ \\ \midrule[1pt]
\multirow{6}{*}{\makecell[c]{German}} & MI & 0.716 $\pm$ 0.014 & 0.719 $\pm$ 0.025 & 0.693 $\pm$ 0.044 & 0.635 $\pm$ 0.060 \\
 & GEOM & 0.723 $\pm$ 0.028 & 0.704 $\pm$ 0.031 & 0.695 $\pm$ 0.024 & 0.679 $\pm$ 0.029 \\
 & KARMA & 0.731 $\pm$ 0.019 & 0.723 $\pm$ 0.019 & 0.706 $\pm$ 0.036 & \textbf{0.714} $\pm$ \textbf{0.025} \\
 & genRBF & \textbf{0.747} $\pm$ \textbf{0.011} & \textbf{0.743} $\pm$ \textbf{0.021} & 0.705 $\pm$ 0.026 & 0.687 $\pm$ 0.045\\
 & NewImp & 0.740 $\pm$ 0.014 & 0.737 $\pm$ 0.025 & 0.708 $\pm$ 0.029 & 0.705 $\pm$ 0.026\\
 & Ours & 0.743 $\pm$ 0.024 & 0.733 $\pm$ 0.021 & \textbf{0.723} $\pm$ \textbf{0.024} $\bullet$ & 0.710 $\pm$ 0.019 \\ \midrule[1pt]
\multirow{6}{*}{\makecell[c]{Heart}} & MI & 0.813 $\pm$ 0.029 & 0.801 $\pm$ 0.033 & 0.762 $\pm$ 0.046 & 0.724 $\pm$ 0.040 \\
 & GEOM & 0.806 $\pm$ 0.021 & 0.752 $\pm$ 0.073 & 0.758 $\pm$ 0.053 & 0.679 $\pm$ 0.077 \\
 & KARMA & 0.784 $\pm$ 0.040 & 0.755 $\pm$ 0.038 & 0.756 $\pm$ 0.031 & 0.687 $\pm$ 0.083\\
 & genRBF & 0.813 $\pm$ 0.031 & 0.779 $\pm$ 0.087 & 0.737 $\pm$ 0.032 & 0.732 $\pm$ 0.037\\
 & NewImp & \textbf{0.827} $\pm$ \textbf{0.022} & 0.806 $\pm$ 0.040 & 0.793 $\pm$ 0.044 & 0.741 $\pm$ 0.048\\
 & Ours & 0.821 $\pm$ 0.033 & \textbf{0.816} $\pm$ \textbf{0.026} $\bullet$ & \textbf{0.813} $\pm$ \textbf{0.027} $\bullet$ & \textbf{0.806} $\pm$ \textbf{0.020} $\bullet$ \\ \midrule[1pt]
\multirow{6}{*}{\makecell[c]{Pima}} & MI & 0.751 $\pm$ 0.021 & 0.735 $\pm$ 0.027 & 0.710 $\pm$ 0.023 & 0.642 $\pm$ 0.046 \\
 & GEOM & 0.721 $\pm$ 0.031 & 0.695 $\pm$ 0.028 & 0.680 $\pm$ 0.058 & 0.666 $\pm$ 0.042 \\
 & KARMA & 0.747 $\pm$ 0.034 & 0.693 $\pm$ 0.026 & 0.668 $\pm$ 0.034 & 0.648 $\pm$ 0.036 \\
 & genRBF & \textbf{0.781} $\pm$ \textbf{0.011} & 0.755 $\pm$ 0.017 & 0.725 $\pm$ 0.022 & 0.659 $\pm$ 0.029\\
 & NewImp & 0.762 $\pm$ 0.012 & 0.749 $\pm$ 0.023 & 0.729 $\pm$ 0.022 & 0.689 $\pm$ 0.030\\
 & Ours & 0.756 $\pm$ 0.026 & \textbf{0.759} $\pm$ \textbf{0.022} & \textbf{0.744} $\pm$ \textbf{0.027} $\bullet$ & \textbf{0.709} $\pm$ \textbf{0.037} \\ 
\bottomrule[1.5pt]
\end{tabular}
}
\end{table*}

\subsubsection{Classification Results on Real-World datasets}
\label{sec: 4.2.3}

In this section, we compared different data imputation approaches using seven real-world datasets. Since all imputation algorithms ultimately serve subsequent tasks, we evaluated their performance by calculating the mean and variance of their classification accuracy on the test data. 

We first mimicked the missing completely at random (MCAR) mechanism by randomly removing some values on the four complete datasets, and the results were presented in Table~\ref{Tab: real-world results}.
For situations with a relatively low missing data ratio, such as $m=20\%$, the differences between the methods are not significant. Even using a simple method like MI can achieve decent predictive performance. 
When $m=40\%$, our proposed method achieves the highest average accuracy and the lowest standard deviation on the \textit{australian}, \textit{heart}, and \textit{pima} datasets, demonstrating the superior stability of our method. On the \textit{german} dataset, our algorithm's performance is only slightly lower than genRBF and NewImp. 
When dealing with high missing data ratios, the prediction accuracy of the GEOM, KARMA, and genRBF methods fluctuates across different datasets. However, our method demonstrates even more significant advancements in such scenarios. For instance, at $m=60\%$, our approach outperforms the second-ranking method by an additional accuracy improvement of approximately $1.7\%$. This significant improvement has also been confirmed through a paired t-test conducted at the 5\% significance level. This performance is further reflected in the case of $m=80\%$, where our algorithm achieves precise and stable classification tasks. 

\begin{table}[tb]
\caption{Comparison of classification accuracy (mean $\pm$ std) with a baseline and three state-of-the-art methods on two real datasets with missing values. The best performance is highlighted in bold.}
\label{Tab: newresults}
\centering
\renewcommand{\arraystretch}{1}
\begin{tabular}{ccc}
\toprule[1.5pt]
\multirow{2.5}{*}{Methods} & \multicolumn{2}{c}{Datasets} \\ \cmidrule[0.75pt]{2-3} 
 & Cylinder $(m\approx10\%)$ & Horse $(m\approx30\%)$ \\ \midrule[1pt]
MI & 0.648 $\pm$ 0.014 & 0.844 $\pm$ 0.027 \\
GEOM & 0.620 $\pm$ 0.021 & 0.848 $\pm$ 0.021 \\
KARMA & \textbf{0.678} $\pm$ \textbf{0.018} & 0.842 $\pm$ 0.020 \\
GenRBF & 0.620 $\pm$ 0.021 & 0.836 $\pm$ 0.028 \\
NewImp & 0.657 $\pm$ 0.013 & 0.854 $\pm$ 0.021 \\
Ours & 0.672 $\pm$ 0.020 & \textbf{0.862} $\pm$ \textbf{0.029} \\ \bottomrule[1.5pt]
\end{tabular}
\end{table}

Next, we evaluated the performance of each method on two datasets with missing values. In these two datasets, the missing values are not manually removed and the data collection process is not known to the user, so the missing mechanism may not be MCAR. Moreover, roughly speaking, the missing values tend to be more concentrated in a few features. The results are presented in Table~\ref{Tab: newresults}. 
For the \textit{cylinder} dataset, predicting for this dataset is challenging for all algorithms. However, our algorithm demonstrates comparable performance to KARMA and significantly outperforms other methods. For the \textit{horse} dataset, the proportion of missing values is even higher, which provides our algorithm with greater freedom to learn the relationships between the samples. According to the paired t-test at the 5\% significance level, our algorithm statistically performs better than other methods in multiple experiments.

In addition, some studies have evaluated imputation performance by training multi-layer perceptrons (MLPs) on imputed data \citep{le2021sa, morvan2025imputation}. To test beyond kernel-based methods and SVM, we followed the experimental setup in \citep{le2021sa} and selected MI, NewImp, and our method for comparison (as GEOM, KARMA, and genRBF are only applicable to kernel machines). For the previously tested datasets \textit{australian}, \textit{german}, \textit{heart}, and \textit{pima}, we set the missing rate to $m=50\%$ and reported the results of imputation followed by MLP in Table~\ref{Tab: DI_MLP}. It can be observed that our method still achieves the best classification accuracy in the vast majority of cases and produces more stable results in multiple experiments on the \textit{australian} and \textit{pima} datasets. This demonstrates that our label-guided imputation method not only performs well with SVM but also extends its advantages to other classifiers.

\begin{table*}[t]
\caption{Comparison of classification accuracy (mean $\pm$ std) of different data imputation methods + MLP. The best performance is highlighted in bold.}
\label{Tab: DI_MLP}
\centering
\begin{tabular}{ccccc}
\toprule[1.5pt]
 \multirow{2.5}{*}{\makecell[c]{Methods / \\ datasets}} & \multicolumn{4}{c}{$m=50\%$} \\ \cmidrule[0.75pt]{2-5} 
 & Australian & German & Heart & Pima \\ \midrule[1pt]
 MI & 0.854 $\pm$ 0.025 & 0.710 $\pm$ 0.026 & 0.792 $\pm$ 0.031 & 0.721 $\pm$ 0.022 \\
 NewImp & 0.857 $\pm$ 0.020 & \textbf{0.732} $\pm$ \textbf{0.026} & 0.796 $\pm$ 0.028 & 0.729 $\pm$ 0.033\\
 Ours & \textbf{0.866} $\pm$ \textbf{0.017} & 0.723 $\pm$ 0.033 & \textbf{0.806} $\pm$ \textbf{0.038} & \textbf{0.740} $\pm$ \textbf{0.018} \\
\bottomrule[1.5pt]
\end{tabular}
\end{table*}

\begin{figure}[t]
    \centering
    \includegraphics[width=0.8\linewidth]{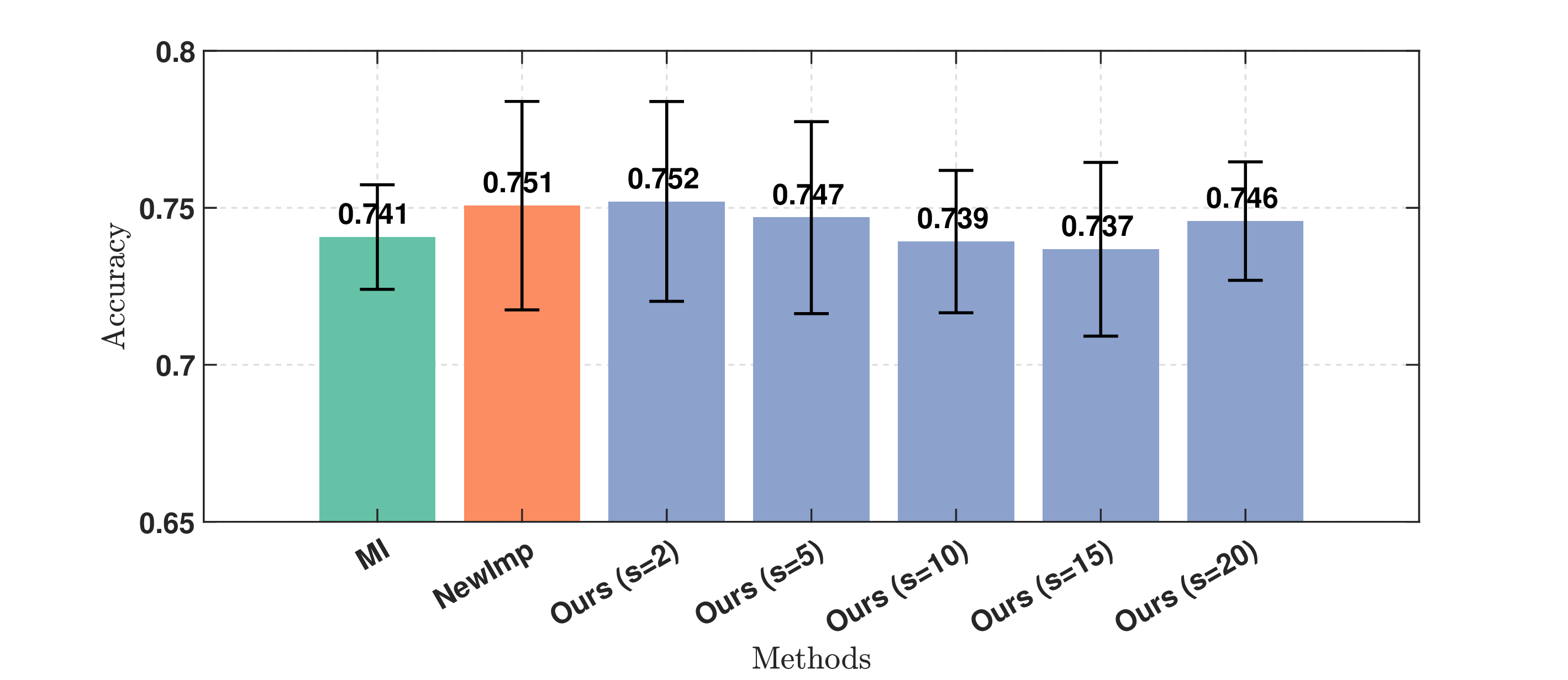}
	\caption{Classification accuracy (mean $\pm$ std) on the \textit{wine} dataset across different splits (denoted as $s$) for our method with two baselines (MI and NewImp).}
    \label{Fig: largescale}
\end{figure}

Finally, in the previous experiments, the scale of the datasets remained small to moderate. Since our algorithm focuses on leveraging pairwise information between data points to guide the imputation process, handling an $N \times N$ matrix is unavoidable, which makes it challenging to directly scale to large datasets. To address this limitation, we split the dataset into subsets and applied our algorithm to each subset to reduce computational costs. The imputation results from each subset were then combined. Subsequently, we compared the imputation results obtained by MI, NewImp, and our method under different dataset splits and applied them to downstream MLP classification tasks to evaluate performance. The results are presented in Figure~\ref{Fig: largescale}. As shown in the results, the accuracy of our method for downstream tasks does exhibit a slight decline as the number of subsets increases. However, when the number of splits is small (e.g., 2 or 5), it still performs comparably to NewImp.

\begin{table*}[t]
\caption{Comparison of inter-class distance (ICD), Fisher discriminant ratio (FDR), Calinski-Harabasz index (CHI), and Davies-Bouldin index (DBI) values of different imputation methods on different datasets. $(\uparrow)$ means that we expect the value to be larger, $(\downarrow)$ vice versa.}
\label{Tab: Metics}
\centering
\begin{tabular}{cccccc}
\toprule[1.5pt]
\multirow{2.5}{*}{\makecell[c]{datasets}} & \multirow{2.5}{*}{\makecell[c]{{Methods}}} & \multicolumn{4}{c}{Metrics} \\ \cmidrule[0.75pt]{3-6} 
 &  & ICD$(\uparrow)$ & FDR$(\uparrow)$ & CHI$(\uparrow)$ & DBI$(\downarrow)$ \\ \midrule[1pt]
\multirow{3}{*}{\makecell[c]{Australian}} & MI & 0.4329 & 0.1608 & 21.655 & 3.416 \\
 & NewImp & 0.4363 & 0.1592 & 21.411 & 3.407 \\
 & Ours & 0.4398 & 0.1666 & 22.416 & 3.365 \\ \midrule[1pt]
\multirow{3}{*}{\makecell[c]{German}} & MI & 0.2603 & 0.0263 & 4.480 & 8.537 \\
 & NewImp & 0.2762 & 0.0280 & 4.781 & 8.243 \\
 & Ours & 0.3873 & 0.0647 & 10.989 & 6.658 \\ \midrule[1pt]
\multirow{3}{*}{\makecell[c]{Heart}} & MI & 0.4455 & 0.1481 & 7.738 & 3.619 \\
 & NewImp & 0.5196 & 0.1963 & 10.280 & 3.207 \\
 & Ours & 0.5730 & 0.2745 & 14.356 & 3.062 \\ \midrule[1pt]
\multirow{3}{*}{\makecell[c]{Pima}} & MI & 0.1112 & 0.0592 & 8.597 & 5.249 \\
 & NewImp & 0.1022 & 0.0510 & 7.426 & 6.037 \\
 & Ours & 0.1180 & 0.0660 & 9.495 & 5.042 \\ 
\bottomrule[1.5pt]
\end{tabular}
\end{table*}

\subsection{Structural Differences of Different Data Imputation Methods}\label{sec: 4.3}

\begin{figure}[t]
    \centering
    \begin{subfigure}{0.32\textwidth}
        \includegraphics[width=\textwidth]{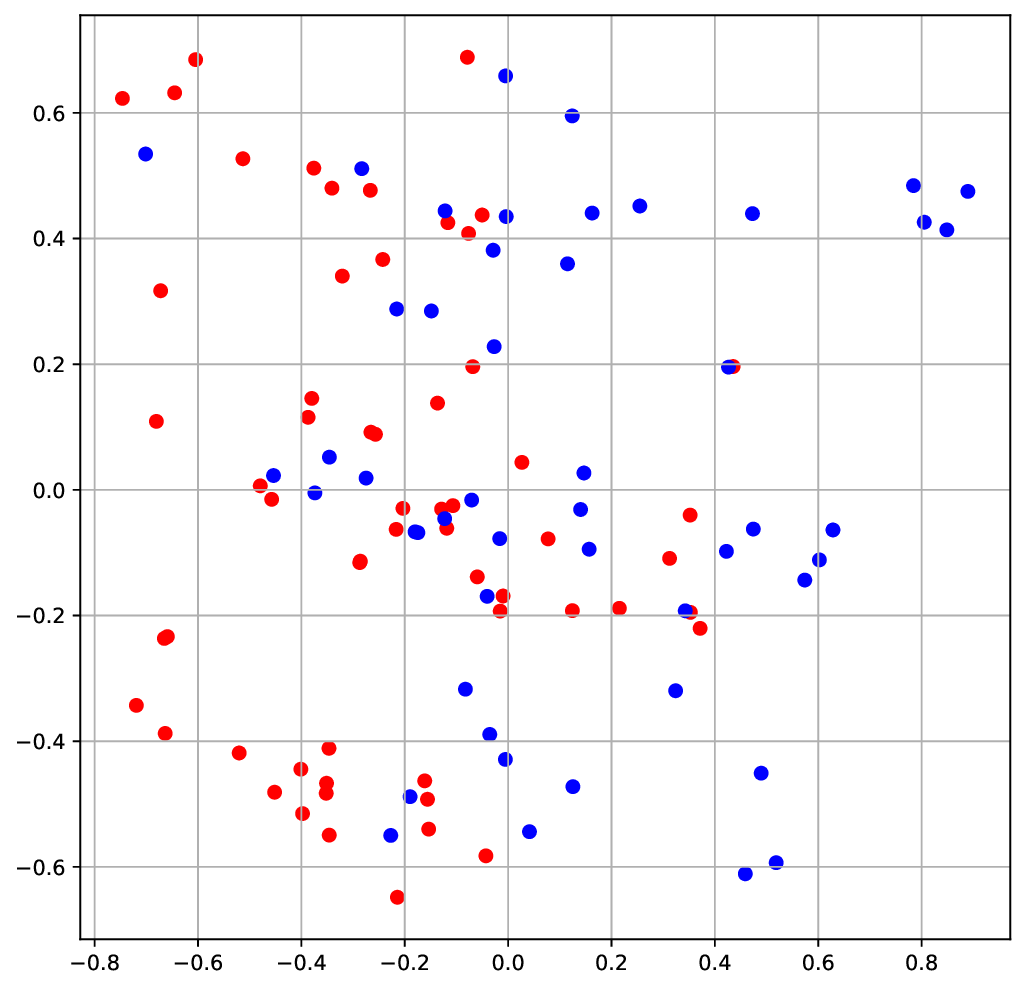}
        \caption{MI}
        \label{fig:heart_MI}
    \end{subfigure}
    \hfill
    \begin{subfigure}{0.32\textwidth}
        \includegraphics[width=\textwidth]{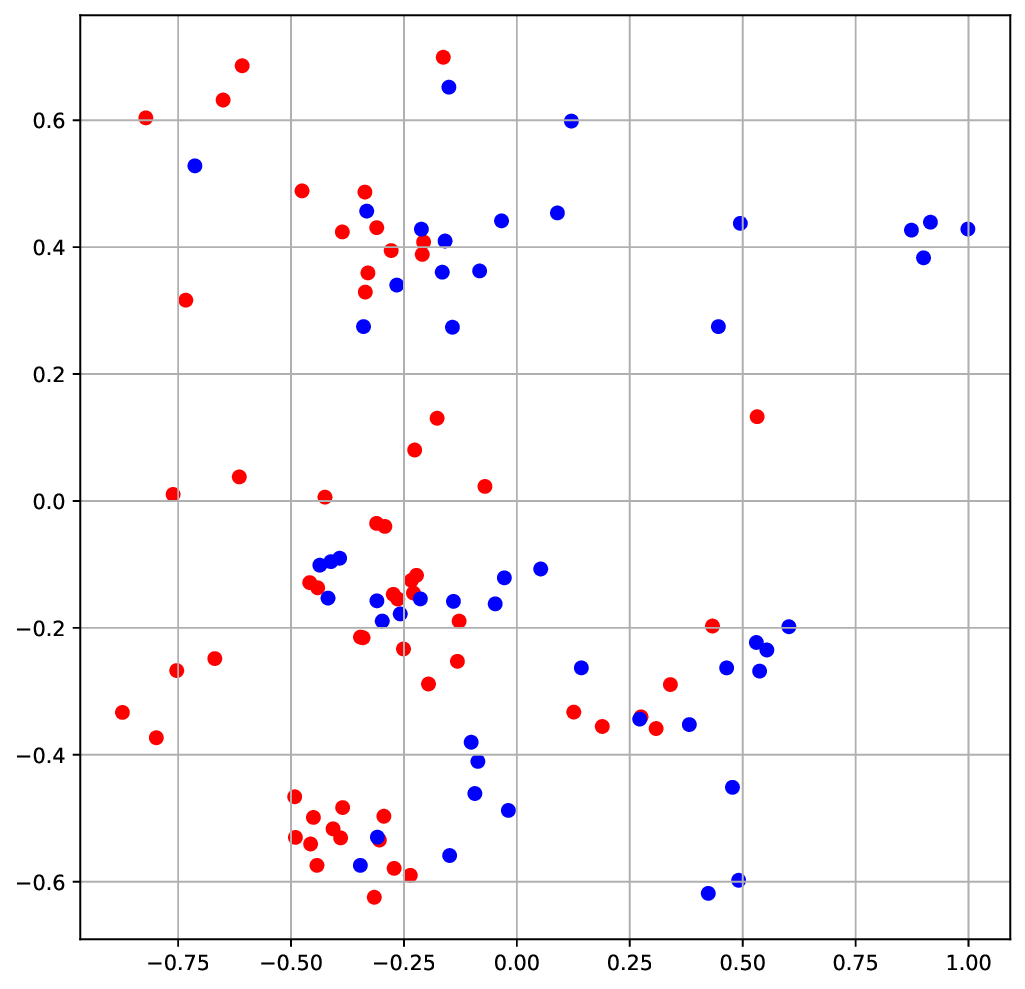}
        \caption{NewImp}
        \label{fig:heart_NewImp}
    \end{subfigure}
    \hfill
    \begin{subfigure}{0.32\textwidth}
        \includegraphics[width=\textwidth]{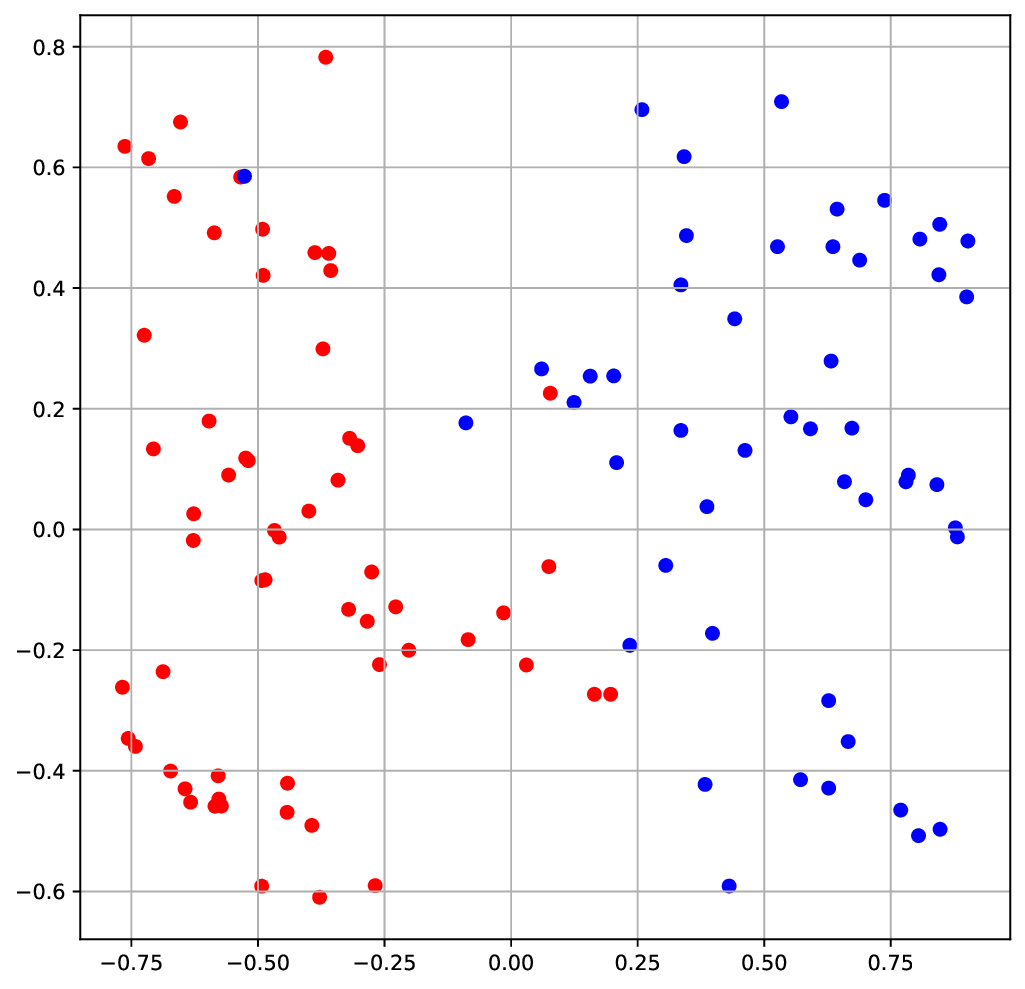}
        \caption{Ours}
        \label{fig:heart_Ours}
    \end{subfigure}

    \caption{PCA visualization of different imputation methods on \textit{heart} dataset.}
    \label{fig:comparison}
\end{figure}

In this section, we explore the structural differences between various imputation methods from both numerical and visual perspectives. Numerically, we employ four clustering metrics: inter-class distance (ICD), Fisher discriminant ratio (FDR), Calinski-Harabasz index (CHI), and Davies-Bouldin index (DBI), to measure whether the imputed data tends to overlap or remains relatively well-separated, which could affect subsequent classification tasks. The definitions of these metrics are provided in~\ref{secD}, and the results are summarized in Table~\ref{Tab: Metics}. From the results, it can be observed that our method achieves larger inter-class distances and better distinguishability in multiple scenarios, with more significant differences from MI and NewImp on the \textit{german} and \textit{heart} datasets. These differences can be more intuitively observed in Figure~\ref{fig:comparison}, where we present the results on the \textit{heart} dataset and perform principal component analysis (PCA) dimensionality reduction visualizations for each method. Specifically, to ensure fairness, we used the PCA projection matrix obtained from the complete data to project the imputed data of MI, NewImp, and our method. Clearly, the samples imputed by MI and NewImp remain overlapped, whereas the proposed method demonstrates more distinct separation between different classes in the imputed two-dimensional subspace. This improved separation is expected to be more beneficial for subsequent classification tasks.

\section{Conclusion}
This paper proposed a novel two-stage data imputation framework, aiming to optimize the similarity relationships between data to guide the completion of missing features by pursuing better classification accuracy. In the first stage, we unify the tasks of kernel matrix imputation and classification within a single framework, enabling mutual guidance between the two tasks in an alternating optimization process to improve similarity relationships. The introduction of a perturbation variable enhances the robustness of the algorithm's predictions. In the second stage, we achieve, for the first time, the recovery of data features from a given kernel matrix, effectively utilizing the optimized information obtained in the first stage. By leveraging the supervision information through two stages, we have obtained a more flexible approach for data imputation, which provides significant advantages when dealing with high missing rates. Numerical experiments have validated that our algorithm achieves higher prediction accuracy and demonstrates more stable performance compared to other methods. In future research, extending the ideas of this study to neural networks would be a fascinating endeavor. This expansion would provide us with a way to handle even larger-scale data. It is important to note that neural networks typically operate in over-parameterized regimes, leveraging their high model complexity to estimate labels effectively. However, this also makes it challenging to control the impact of supervised information on the results of data imputation in the absence of appropriate regularization terms.

\section*{Acknowledgment}
This work was supported by the National Natural Science Foundation of China (62376155). The authors would like to thank the anonymous AE and reviewers for their insightful comments.

\bibliographystyle{elsarticle-num} 
\bibliography{ref}

\appendix
\section{Decoupling of Various Kernel Function Computations}
\label{secA}

Below, we have provided a list of commonly used kernel functions that are decoupled. By separating the observed and unknown components, we compute the observed kernel matrix and establish an upper limit for the adjustment matrix. This approach allows us to fully leverage all the observed features in the data (assuming $\vx\in[0,1]^d$). Recall that an incomplete dataset is defined as $\mathcal{D}=\lbrace \vx_{\vo_i}^i, y_i\rbrace _{i=1}^N$, where $\vo_i$ represents the set of observable features, $\vx_{\vo_i}^i\in(\mathbb{R}\cup \{\ast\})^d$, and $\ast$ represents an unknown real number within the range $[0, 1]$. For the following kernel functions, the computation formula for the $(i,j)$-th element of the kernel matrix $\mK$ can be represented as:
\begin{enumerate}
    \item Linear Kernel: $k(\vx,\vy)=\vx^\top\vy$, and 
        \begin{equation*}
            \begin{aligned}
                &K_{i,j}=\left(\vx_{\vo_i}^i\right)^\top \vx_{\vo_j}^j\\
                &=\sum_{p\in\vo_i\cap\vo_j}\left(x_p^i\cdot x_p^j\right)+\sum_{p\in\vo_i\backslash\vo_j}\left(x_p^i\cdot\ast\right)+\sum_{p\in\vo_j\backslash\vo_i}\left(\ast\cdot x_p^j\right)+\sum_{p\notin\vo_i\cup\vo_j}\left(\ast\cdot\ast\right).
            \end{aligned}
        \end{equation*}
    \item Polynomial Kernel: $k(\vx,\vy)=\left(\vx^\top\vy+r\right)^d\ (r\geq 0\ \mathrm{and}\ d\geq 1)$, and
        \begin{equation*}
            \begin{aligned}
                &K_{i,j}=\left(D_{i,j}+r\right)^d \ \mathrm{where}\\
                &D_{i,j}=\left(\vx_{\vo_i}^i\right)^\top \vx_{\vo_j}^j\\
                &=\sum_{p\in\vo_i\cap\vo_j}\left(x_p^i\cdot x_p^j\right)+\sum_{p\in\vo_i\backslash\vo_j}\left(x_p^i\cdot\ast\right)+\sum_{p\in\vo_j\backslash\vo_i}\left(\ast\cdot x_p^j\right)+\sum_{p\notin\vo_i\cup\vo_j}\left(\ast\cdot\ast\right).
            \end{aligned}
        \end{equation*}
    \item Gaussian Kernel: $k(\vx,\vy)=\mathrm{exp}\left(-\gamma\|\vx-\vy\|_2^2\right)$, and
        \begin{equation*}
            \begin{aligned}
                &K_{i,j}=\mathrm{exp}\left(-\gamma D_{i,j}\right) \ \mathrm{where}\\
                &D_{i,j}=\|\vx_{\vo_i}^i- \vx_{\vo_j}^j\|_2^2\\
                &=\sum_{p\in\vo_i\cap\vo_j}\left(x_p^i-x_p^j\right)^2+\sum_{p\in\vo_i\backslash\vo_j}\left(x_p^i-\ast\right)^2+\sum_{p\in\vo_j\backslash\vo_i}\left(\ast-x_p^j\right)^2+\sum_{p\notin\vo_i\cup\vo_j}\left(\ast-\ast\right)^2.
            \end{aligned}
        \end{equation*}
    \item Laplacian Kernel: $k(\vx,\vy)=\mathrm{exp}\left(-\gamma\|\vx-\vy\|_1\right)$, and 
        \begin{equation*}
            \begin{aligned}
                &K_{i,j}=\mathrm{exp}\left(-\gamma D_{i,j}\right) \ \mathrm{where}\\
                &D_{i,j}=\|\vx_{\vo_i}^i- \vx_{\vo_j}^j\|_1\\
                &=\sum_{p\in\vo_i\cap\vo_j}|x_p^i-x_p^j|+\sum_{p\in\vo_i\backslash\vo_j}|x_p^i-\ast|+\sum_{p\in\vo_j\backslash\vo_i}|\ast-x_p^j|+\sum_{p\notin\vo_i\cup\vo_j}|\ast-\ast|.
            \end{aligned}
        \end{equation*}
    \item Sigmoid Kernel: $k(\vx,\vy)=\mathrm{tanh}\left(\gamma\vx^\top\vy+r\right)\ (\gamma > 0\ \mathrm{and}\ r\geq 0)$, and 
        \begin{equation*}
            \begin{aligned}
                &K_{i,j}=\mathrm{tanh}\left(\gamma D_{i,j}+r\right) \ \mathrm{where}\\
                &D_{i,j}=\left(\vx_{\vo_i}^i\right)^\top \vx_{\vo_j}^j\\
                &=\sum_{p\in\vo_i\cap\vo_j}\left(x_p^i\cdot x_p^j\right)+\sum_{p\in\vo_i\backslash\vo_j}\left(x_p^i\cdot\ast\right)+\sum_{p\in\vo_j\backslash\vo_i}\left(\ast\cdot x_p^j\right)+\sum_{p\notin\vo_i\cup\vo_j}\left(\ast\cdot\ast\right).
            \end{aligned}
        \end{equation*}
    \item $\chi^2$ Kernel: $k(\vx,\vy)=\sum_{p=1}^d\frac{2x_py_p}{x_p+y_p}$, and
        \begin{equation*}
            \begin{aligned}
                &K_{i,j}=\sum_{p=1}^d\frac{2x_p^ix_p^j}{x_p^i+x_p^j}\\
                &=\sum_{p\in\vo_i\cap\vo_j}\frac{2x_p^ix_p^j}{x_p^i+x_p^j}+\sum_{p\in\vo_i\backslash\vo_j}\frac{2(x_p^i\cdot\ast)}{x_p^i+\ast}+\sum_{p\in\vo_j\backslash\vo_i}\frac{2(\ast \cdot x_p^j)}{\ast+x_p^j}+\sum_{p\notin\vo_i\cup\vo_j}\frac{2(\ast\cdot\ast)}{\ast+\ast}.
            \end{aligned}
        \end{equation*}
\end{enumerate}
The range of the second term depends on $x_p^i$, while the range of the third term depends on $x_p^j$. The range of the fourth term is $[0, d-|\vo_i\cup\vo_j|]$. Moreover, these kernel functions can all be adapted to the algorithm utilized in Stage II of our framework.

\section{Convergence Proof}
\label{secB}
To demonstrate the convergence of the three-level optimization in 
Stage I, we first review the optimization process at each step and list relevant parameter descriptions.

For the three optimization steps in Stage I, they are equivalent to $\mK_\Delta \coloneqq \mathcal{P}_1(\hat{\mK}_\Delta)$, $\mathcal{E} \coloneqq \mathcal{P}_2(\hat{\mathcal{E}})$, and $\valpha \coloneqq \mathcal{P}_3(\hat{\valpha})$, where $\hat{\mK}_\Delta$, $\hat{\mathcal{E}}$, and $\hat{\valpha}$ are obtained through closed-form solutions and gradient descent ($t$ is the iteration number): 
\begin{equation}
    \mathcal{L}=\vone^\top\valpha-\frac{1}{2}\valpha^\top\diag(\vy)\left(\mK_o\odot\mK_{\Delta}\odot\mathcal{E}\right)\diag(\vy)\valpha+\eta\|\mK_{\Delta}-\vone\vone^\top\|_{\mathrm{F}}^2
\end{equation}
\begin{itemize}
    \item Step 1.
    \begin{subequations}
        \begin{align}
            \hat{\mK}_{\Delta}^{(t)}& = \vone\vone^\top+\frac{1}{4\eta}\diag(\valpha^{(t-1)}\odot\vy) \left(\mK_o\odot \mathcal{E}^{(t-1)}\right)\diag(\valpha^{(t-1)}\odot\vy), \\
            \mK_\Delta^{(t)} &= \mathcal{P}_1(\hat{\mK}_{\Delta}^{(t)}) = \argmin_{\mK\in\mathcal{C}_1}\  \left\|\mK-\hat{\mK}_{\Delta}^{(t)}\right\|_{\mathrm{F}}^2,\quad
            \mathcal{C}_1 = \{\mK\mid\mB_l\preceq \mK \preceq \mB_u,\  \mK\in\mathcal{S}_{+}\}.\label{equ:s1}
        \end{align}
    \end{subequations}
    \item Step 2.
    \begin{subequations}
        \begin{align}
            \hat{\mathcal{E}}^{(t)} &= \vone\vone^\top-\frac{1}{4\rho}\diag(\valpha^{(t-1)}\odot\vy) \left(\mK_o\odot \mK_\Delta^{(t)}\right)\diag(\valpha^{(t-1)}\odot\vy), \\
            \mathcal{E}^{(t)} &=\mathcal{P}_2(\hat{\mathcal{E}}^{(t)}) = \argmin_{\mathcal{E}\in\mathcal{C}_2}\  \left\|\mathcal{E}-\hat{\mathcal{E}}^{(t)}\right\|_{\mathrm{F}}^2,\quad\quad
            \mathcal{C}_2 =\{\mathcal{E}\mid\mathcal{E}\in\mathcal{S}_{+}\}.\label{equ:s2}
        \end{align}
    \end{subequations}
    \item Step 3.
    \begin{subequations}
        \begin{align}
            \hat{\valpha}^{(t)} &= \hat{\valpha}^{(t-1)} +  \nu^{(t)} \nabla_{\valpha}  \mathcal{L}(\hat{\valpha}^{(t-1)};\mK_\Delta^{(t)},\mathcal{E}) \\ 
            \valpha^{(t)} &= \mathcal{P}_3(\hat{\valpha}) = \argmin_{\valpha\in\mathcal{C}_3}\  \left\|\valpha-\hat{\valpha}^{(t)}\right\|_{2}^2,\quad\quad
            \mathcal{C}_3 =\{\vy^\top\valpha=0 , \ \vzero\leq \valpha\leq C\vone\}.\label{equ:s3}
        \end{align}
    \end{subequations}
\end{itemize}

\textbf{Notations.} $N$ is the training sample size. $C$ is the hyperparameter of the SVM. In our algorithm, $\eta$, $\rho$, and $\nu^{(t)}$ control the optimization step sizes for variables $\mK_\Delta$, $\mathcal{E}$, and $\valpha$ in each iteration, respectively. Here, $\eta$ and $\rho$ are predetermined, while $\nu^{(t)}$ decreases gradually with iterations. Additionally, we make the following assumptions:
\begin{assumption}\label{assp:1}
    The kernel is bounded, and the norm of the kernel matrix is also bounded, which can be expressed as $\|\mK_o\|_\mathrm{F} \leq \kappa$, $\|\mK_o\|_2 \leq \lambda$. Meanwhile, we set $b_\mathrm{min} \leq \min\{\mB_l\} \leq \max\{\mB_u\} \leq b_\mathrm{max}$.
\end{assumption}
Note that for commonly used RBF kernels, such as the Gaussian kernel, it also holds that $\lambda \leq N$ and $0 \leq b_\mathrm{min} \leq b_\mathrm{max} \leq 1$.

Since the sets $\mathcal{C}_i$ ($i \in \{1, 2, 3\}$) spanned by the constraints in equation~(\ref{equ:s1}), (\ref{equ:s2}), (\ref{equ:s3}) are all closed, non-empty, and convex, $\mathcal{P}_i$ ($i \in \{1, 2, 3\}$) are all projection operators. Thus, they are firmly non-expansive and have the following property:
\begin{property}\label{proper:1}
    For $\mathcal{P}_i$ and $\forall \mX_1, \mX_2, \mX \in\mathcal{C}_i$ ($i \in \{1, 2, 3\}$) along with an arbitrary matrix or vector $\mM$, it satisfies 
    \begin{enumerate}
        \item $\|\mathcal{P}_i(\mX_1) - \mathcal{P}_i(\mX_2)\|_{\mathrm{F}\  \mathrm{or}\ 2} \leq \|\mX_1 - \mX_2\|_{\mathrm{F}\  \mathrm{or}\ 2}$.
        \item $\|\mathcal{P}_i(\mM) - \mM\|_{\mathrm{F}\  \mathrm{or}\ 2}\leq\|\mX-\mM\|_{\mathrm{F}\  \mathrm{or}\ 2}$.
    \end{enumerate}
\end{property}
Here, the first condition follows from the definition of non-expansive operators, while the second results from the fact that for any matrix or vector $\mM$, its projection onto $\mathcal{C}_i$ necessarily has minimal distance to other elements in $\mathcal{C}_i$.
\begin{lemma}\label{lemma:2}
    Recalling the notation from Assumption~\ref{assp:1}, the following inequality holds for the norm of the Hadamard product between $\mK_o$ and $\mathcal{E}$:
    \begin{subequations}
        \begin{align}
            \|\mK_o\odot\mathcal{E}\|_\mathrm{F}&\leq\kappa\left(1+\frac{NC^2\kappa b_{\mathrm{max}}}{4\rho}\right),\label{equ:lemma2.1}\\
            \|\mK_o\odot\mathcal{E}\|_2&\leq\kappa\left(1+\frac{NC^2\lambda b_{\mathrm{max}}}{4\rho}\right).\label{equ:lemma2.2}
        \end{align}
    \end{subequations}
\end{lemma}
\noindent\textit{Proof of Lemma~\ref{lemma:2}.}
\begin{equation}\label{equ: B2}
    \begin{aligned}
        \|\mK_o\odot\mathcal{E}\|_\mathrm{F} &= \|\mK_o\odot(\vone\vone^\top+\mathcal{E}-\vone\vone^\top)\|_\mathrm{F} \\
        &= \|\mK_o + \mK_o\odot(\mathcal{E}-\vone\vone^\top)\|_\mathrm{F} \\
        &\overset{(a)}{\leq} \|\mK_o\|_\mathrm{F} + \|\mK_o\odot(\mathcal{E}-\vone\vone^\top)\|_\mathrm{F} \\
        &\overset{(b)}{\leq} \kappa(1+\|\mathcal{E}-\vone\vone^\top\|_\mathrm{F}).
    \end{aligned}
\end{equation}
Here, inequality (a) is obtained through the triangle inequality of norms, while inequality (b) uses the fact that $\|\mA\odot\mB\|_\mathrm{F}\leq \|\mA\|_\mathrm{F}\max_{i,j}|\mB_{i,j}|\leq\|\mA\|_\mathrm{F}\|\mB\|_\mathrm{F}$. Note that $\mathcal{E}$ here is the result after projection $\mathcal{P}_2(\hat{\mathcal{E}})$. From the second inequality in Property~\ref{proper:1}, substituting $\mM = \hat{\mathcal{E}} = \vone\vone^\top - \frac{1}{4\rho}\diag(\valpha\odot\vy) \left(\mK_o\odot \mK_\Delta\right)\diag(\valpha\odot\vy)$ and $\mX = \vone\vone^\top$, we have:
\begin{equation}\label{equ: B3}
    \begin{aligned}
        \|\mathcal{E}-\vone\vone^\top\|_\mathrm{F} &\overset{(a)}{=} \left\|\mathcal{P}_2\left(\vone\vone^\top - \frac{1}{4\rho}\diag(\valpha\odot\vy) \left(\mK_o\odot \mK_\Delta\right)\diag(\valpha\odot\vy)\right)-\mathcal{P}_2(\vone\vone^\top)\right\|_\mathrm{F} \\
        & \leq \left\|\vone\vone^\top - \frac{1}{4\rho}\diag(\valpha\odot\vy) \left(\mK_o\odot \mK_\Delta\right)\diag(\valpha\odot\vy)-\vone\vone^\top\right\|_\mathrm{F} \\
        & = \frac{1}{4\rho}\left\|\diag(\valpha\odot\vy) \left(\mK_o\odot \mK_\Delta\right)\diag(\valpha\odot\vy)\right\|_\mathrm{F} \\
        &\leq \frac{1}{4\rho}\left\|\diag(\valpha\odot\vy)\|_\mathrm{F}^2 \|\mK_o\odot \mK_\Delta\right\|_\mathrm{F} \\
        &\overset{(b)}{\leq} \frac{NC^2\kappa b_{\mathrm{max}}}{4\rho}.
    \end{aligned}
\end{equation}
Equality (a) holds because $\vone\vone^\top$ is PSD, implying \(\vone\vone^\top = \mathcal{P}_2(\vone\vone^\top)\). For inequality (b), it follows from $\vzero \leq \alpha \leq C\vone$ and $\mB_l \preceq \mK_\Delta \preceq \mB_u$, which lead to $\|\diag(\valpha \odot \vy)\|_\mathrm{F}^2 \leq NC^2$ and $\|\mK_o \odot \mK_\Delta\|_\mathrm{F} \leq \|\mK_o\|_\mathrm{F} \cdot\max\{\mK_\Delta\}\leq\kappa b_{\mathrm{max}}$. Then, substituting equation~(\ref{equ: B3}) into equation~(\ref{equ: B2}) yields $\|\mK_o\odot\mathcal{E}\|_\mathrm{F}\leq\kappa\left(1+\frac{NC^2\kappa b_{\mathrm{max}}}{4\rho}\right)$. Similarly, for the $\|\mK_o\odot\mathcal{E}\|_2$, we can use the fact that $\|\mA\odot\mB\|_2\leq\|\mA\|_2\max_{i,j}|\mB_{i,j}|$ when $\mB$ is PSD. Hence, $\|\mK_o\odot \mK_\Delta\|_2\leq \|\mK_o\|_2\max_{i,j} {(\mK_\Delta)}_{i,j}\leq \lambda b_{\mathrm{max}}$,  which completes the proof.
\qed

\begin{lemma} (Bounds related to $\valpha$-updates) \label{lemma:3}
    For iteration index $t$ and any two consecutive iterates \( \valpha^{(t)} \) and \( \valpha^{(t-1)} \), it holds that
    \begin{equation}
        \begin{aligned}
            & \left\|\diag(\valpha^{(t)}\odot\vy) \left(\mK_o\odot \mathcal{E}\right)\diag(\valpha^{(t)}\odot\vy) - \diag(\valpha^{(t-1)}\odot\vy) \left(\mK_o\odot \mathcal{E}\right)\diag(\valpha^{(t-1)}\odot\vy)\right\|_\mathrm{F} \\
            & \leq 2\nu^{(t)}NC\kappa\left(1+\frac{NC^2\kappa b_{\mathrm{max}}}{4\rho}\right)\left(1+C\kappa b_{\mathrm{max}} + \frac{NC^3\kappa\lambda b_{\mathrm{max}}^2}{4\rho}\right).
        \end{aligned}
    \end{equation}
\end{lemma}
\noindent\textit{Proof of Lemma~\ref{lemma:3}.}
\begin{equation}\label{equ:B.6}
    \begin{aligned}
        & \left\|\diag(\valpha^{(t)}\odot\vy) \left(\mK_o\odot \mathcal{E}\right)\diag(\valpha^{(t)}\odot\vy) - \diag(\valpha^{(t-1)}\odot\vy) \left(\mK_o\odot \mathcal{E}\right)\diag(\valpha^{(t-1)}\odot\vy)\right\|_\mathrm{F} \\
        & =  \left\|\diag(\valpha^{(t)}\odot\vy+\valpha^{(t-1)}\odot\vy) \left(\mK_o\odot \mathcal{E}\right)\diag(\valpha^{(t)}\odot\vy-\valpha^{(t-1)}\odot\vy)\right\|_\mathrm{F} \\
        & \leq \|\mK_o\odot\mathcal{E}\|_\mathrm{F}\|\diag(\valpha^{(t)}\odot\vy+\valpha^{(t-1)}\odot\vy)\|_\mathrm{F}\|\diag(\valpha^{(t)}\odot\vy-\valpha^{(t-1)}\odot\vy)\|_\mathrm{F} \\
        & \overset{(a)}{\leq} \kappa\left(1+\frac{NC^2\kappa b_{\mathrm{max}}}{4\rho}\right) \|\valpha^{(t)}+\valpha^{(t-1)}\|_2\|\valpha^{(t)}-\valpha^{(t-1)}\|_2 \\
        &= \kappa\left(1+\frac{NC^2\kappa b_{\mathrm{max}}}{4\rho}\right) \sqrt{N(2C)^2}\ \|\mathcal{P}_3(\hat{\valpha}^{(t)})-\mathcal{P}_3(\hat{\valpha}^{(t-1)})\|_2 \\
        & \overset{(b)}{\leq} \kappa\left(1+\frac{NC^2\kappa b_{\mathrm{max}}}{4\rho}\right) \sqrt{N(2C)^2}\ \|\hat{\valpha}^{(t)}-\hat{\valpha}^{(t-1)}\|_2.
    \end{aligned}
\end{equation}
Here, inequality (a) is obtained by substituting equation~(\ref{equ:lemma2.1}) from Lemma~\ref{lemma:2}, while inequality (b) utilizes Property~\ref{proper:1}. And the difference between $\hat{\valpha}^{(t)}$ and $\hat{\valpha}^{(t-1)}$ is simply the learning rate $\nu^{(t)}$ multiplied by the gradient, which can be expressed as:
\begin{equation}\label{equ:B.7}
    \begin{aligned}
        \|\hat{\valpha}^{(t)}-\hat{\valpha}^{(t-1)}\|_2 &= \|\nu^{(t)}\nabla_{\valpha}\mathcal{L}(\hat{\valpha}^{(t-1)};\mK_\Delta, \mathcal{E})\|_2 \\
        &= \nu^{(t)}\| \vone- \diag(\vy) \left(\mK_o\odot \mK_\Delta\odot\mathcal{E}\right)\diag(\vy)\hat{\valpha}^{(t-1)}\|_2 \\
        & \leq \nu^{(t)}\left(\sqrt{N}+ \|\mK_o\odot \mK_\Delta\odot\mathcal{E}\|_2\|\hat{\valpha}^{(t-1)}\|_2\right) \\
        &\leq \nu^{(t)}\left(\sqrt{N}+\|\hat{\valpha}^{(t-1)}\|_2 \max_{i,j}(\mK_\Delta)_{i,j}\|\mK_o\odot\mathcal{E}\|_2\right) \\
        & \overset{(a)}{\leq} \nu^{(t)}\left[\sqrt{N}+ \sqrt{N}C \cdot b_{\mathrm{max}}\cdot \kappa\left(1+\frac{NC^2\lambda b_{\mathrm{max}}}{4\rho}\right) \right],
    \end{aligned}
\end{equation}
where inequality (a) follows from the constraints on $\valpha$ and $\mK_\Delta$, combined with the equation~(\ref{equ:lemma2.2}) in Lemma~\ref{lemma:2}. Substituting equation~(\ref{equ:B.7}) back into~(\ref{equ:B.6}), we obtain
\begin{equation}
    \begin{aligned}
        & \left\|\diag(\valpha^{(t)}\odot\vy) \left(\mK_o\odot \mathcal{E}\right)\diag(\valpha^{(t)}\odot\vy) - \diag(\valpha^{(t-1)}\odot\vy) \left(\mK_o\odot \mathcal{E}\right)\diag(\valpha^{(t-1)}\odot\vy)\right\|_\mathrm{F} \\
        & \leq \kappa\left(1+\frac{NC^2\kappa b_{\mathrm{max}}}{4\rho}\right) \sqrt{N(2C)^2}\nu^{(t)}\left[\sqrt{N}+ \sqrt{N}C \cdot b_{\mathrm{max}}\cdot \kappa\left(1+\frac{NC^2\lambda b_{\mathrm{max}}}{4\rho}\right) \right] \\
        & = 2\nu^{(t)}NC\kappa\left(1+\frac{NC^2\kappa b_{\mathrm{max}}}{4\rho}\right)\left(1+C\kappa b_{\mathrm{max}} + \frac{NC^3\kappa\lambda b_{\mathrm{max}}^2}{4\rho}\right),
    \end{aligned}
\end{equation}
thereby obtaining the desired result.
\qed

\begin{lemma} (Bounds related to $\mathcal{E}$-updates) \label{lemma:4}
    Similar to the formulation in Lemma~\ref{lemma:3}, during the update of $\mathcal{E}$ in Step 2 of Stage I, the following value will be bounded by
    \begin{equation}
        \begin{aligned}
            &\left\|\diag(\valpha^{(t)}\odot\vy) \left(\mK_o\odot \mK_\Delta^{(t)}\right)\diag(\valpha^{(t)}\odot\vy) - \diag(\valpha^{(t-1)}\odot\vy) \left(\mK_o\odot \mK_\Delta^{(t-1)}\right)\diag(\valpha^{(t-1)}\odot\vy)\right\|_\mathrm{F} \\
            &\leq NC^2\kappa\left\|\mK_\Delta^{(t)}-\mK_\Delta^{(t-1)}\right\|_\mathrm{F} + 2\nu^{(t)}NC\kappa b_{\mathrm{max}}\left(1+C\kappa b_{\mathrm{max}} + \frac{NC^3\kappa\lambda b_{\mathrm{max}}^2}{4\rho}\right).
        \end{aligned}
    \end{equation}
\end{lemma}
\noindent\textit{Proof of Lemma~\ref{lemma:4}.}
\begin{equation}
    \begin{aligned}
        & \left\|\diag(\valpha^{(t)}\odot\vy) \left(\mK_o\odot \mK_\Delta^{(t)}\right)\diag(\valpha^{(t)}\odot\vy) - \diag(\valpha^{(t-1)}\odot\vy) \left(\mK_o\odot \mK_\Delta^{(t-1)}\right)\diag(\valpha^{(t-1)}\odot\vy)\right\|_\mathrm{F} \\
        & = \left\|\diag(\valpha^{(t)}\odot\vy) \left(\mK_o\odot \mK_\Delta^{(t)}\right)\diag(\valpha^{(t)}\odot\vy) - \diag(\valpha^{(t)}\odot\vy) \left(\mK_o\odot \mK_\Delta^{(t-1)}\right)\diag(\valpha^{(t)}\odot\vy)\right. \\
        &\quad + \left.\diag(\valpha^{(t)}\odot\vy) \left(\mK_o\odot \mK_\Delta^{(t-1)}\right)\diag(\valpha^{(t)}\odot\vy) - \diag(\valpha^{(t-1)}\odot\vy) \left(\mK_o\odot \mK_\Delta^{(t-1)}\right)\diag(\valpha^{(t-1)}\odot\vy)\right\|_\mathrm{F} \\
        & \overset{(a)}{\leq} \left\|\diag(\valpha^{(t)}\odot\vy) \left(\mK_o\odot \mK_\Delta^{(t)}\right)\diag(\valpha^{(t)}\odot\vy) - \diag(\valpha^{(t)}\odot\vy) \left(\mK_o\odot \mK_\Delta^{(t-1)}\right)\diag(\valpha^{(t)}\odot\vy)\right\|_\mathrm{F}\\
        &\quad + \left\|\diag(\valpha^{(t)}\odot\vy) \left(\mK_o\odot \mK_\Delta^{(t-1)}\right)\diag(\valpha^{(t)}\odot\vy) - \diag(\valpha^{(t-1)}\odot\vy) \left(\mK_o\odot \mK_\Delta^{(t-1)}\right)\diag(\valpha^{(t-1)}\odot\vy)\right\|_\mathrm{F} \\
        & \overset{(b)}{=} \|\mathrm{diag}(\valpha^{(t)}\odot \vy)\|_2^2\left\|\mK_o\odot(\mK_\Delta^{(t)}-\mK_\Delta^{(t-1)})\right\|_\mathrm{F} + \|\mK_o\odot\mK_\Delta^{(t-1)}\|_\mathrm{F}\sqrt{N(2C)^2}\|\hat{\valpha}^{(t)}-\hat{\valpha}^{(t-1)}\|_2 \\
        & \overset{(c)}{\leq} NC^2\kappa\left\|\mK_\Delta^{(t)}-\mK_\Delta^{(t-1)}\right\|_\mathrm{F} + \kappa b_{\mathrm{max}}\sqrt{N}2C\nu^{(t)}\left[\sqrt{N}+ \sqrt{N}C \cdot b_{\mathrm{max}}\cdot \kappa\left(1+\frac{NC^2\lambda b_{\mathrm{max}}}{4\rho}\right) \right] \\
        & = NC^2\kappa\left\|\mK_\Delta^{(t)}-\mK_\Delta^{(t-1)}\right\|_\mathrm{F} + 2\nu^{(t)}NC\kappa b_{\mathrm{max}}\left(1+C\kappa b_{\mathrm{max}} + \frac{NC^3\kappa\lambda b_{\mathrm{max}}^2}{4\rho}\right),
    \end{aligned}
\end{equation}
where inequality (a) is obtained via the triangle inequality, the latter term in equation (b) is derived from the proof of equation~(\ref{equ:B.6}), and inequality (c) follows from the result of equation~(\ref{equ:B.7}).

\qed

\begin{theorem} (Bounds related to $\mK_\Delta$-updates) \label{theorem:5}
    For Stage I in Algorithm~\ref{Algo}, the difference in $\mK_\Delta$ between two consecutive updates can be bounded by
    \begin{equation}
        \begin{aligned}
            & \left\|\mK_\Delta^{(t+1)} - \mK_\Delta^{(t)} \right\|_\mathrm{F} \\
            & \leq \frac{1}{4\eta} N^2C^4\kappa^2\left\|\mK_\Delta^{(t)}-\mK_\Delta^{(t-1)}\right\|_\mathrm{F} + \frac{1}{2\eta}\nu^{(t)}N^2C^3\kappa^2 b_{\mathrm{max}}\left(1+C\kappa b_{\mathrm{max}} + \frac{NC^3\kappa\lambda b_{\mathrm{max}}^2}{4\rho}\right)\\
            &\quad + \frac{1}{2\eta} \nu^{(t)}NC\kappa\left(1+\frac{NC^2\kappa b_{\mathrm{max}}}{4\rho}\right)\left(1+C\kappa b_{\mathrm{max}} + \frac{NC^3\kappa\lambda b_{\mathrm{max}}^2}{4\rho}\right)
        \end{aligned}
    \end{equation}
\end{theorem}
\noindent\textit{Proof of Theorem~\ref{theorem:5}.}
Finally, we establish the following result:
\begin{equation}\label{equ:B.12}
    \begin{aligned}
        & \left\|\mK_\Delta^{(t+1)} - \mK_\Delta^{(t)} \right\|_\mathrm{F} = \left\| \mathcal{P}_1(\hat{\mK}_\Delta^{(t+1)}) - \mathcal{P}_1(\hat{\mK}_\Delta^{(t)}) \right\|_\mathrm{F} \overset{(a)}{\leq} \left\| \hat{\mK}_\Delta^{(t+1)} - \hat{\mK}_\Delta^{(t)}\right\|_\mathrm{F} \\
        & \overset{(b)}{=} \frac{1}{4\eta}\left\|\diag(\valpha^{(t)}\odot\vy) \left(\mK_o\odot \mathcal{E}^{(t)}\right)\diag(\valpha^{(t)}\odot\vy) - \diag(\valpha^{(t-1)}\odot\vy) \left(\mK_o\odot \mathcal{E}^{(t-1)}\right)\diag(\valpha^{(t-1)}\odot\vy)\right\|_\mathrm{F} \\
        & = \frac{1}{4\eta}\left\|\diag(\valpha^{(t)}\odot\vy) \left(\mK_o\odot \mathcal{E}^{(t)}\right)\diag(\valpha^{(t)}\odot\vy) - \diag(\valpha^{(t)}\odot\vy) \left(\mK_o\odot \mathcal{E}^{(t-1)}\right)\diag(\valpha^{(t)}\odot\vy)\right.\\
        &\quad + \left.\diag(\valpha^{(t)}\odot\vy) \left(\mK_o\odot \mathcal{E}^{(t-1)}\right)\diag(\valpha^{(t)}\odot\vy) - \diag(\valpha^{(t-1)}\odot\vy) \left(\mK_o\odot \mathcal{E}^{(t-1)}\right)\diag(\valpha^{(t-1)}\odot\vy)\right\|_\mathrm{F}, \\
        & \overset{(c)}{\leq} \frac{1}{4\eta}\underbrace{\left\|\diag(\valpha^{(t)}\odot\vy) \left(\mK_o\odot \mathcal{E}^{(t)}\right)\diag(\valpha^{(t)}\odot\vy) - \diag(\valpha^{(t)}\odot\vy) \left(\mK_o\odot \mathcal{E}^{(t-1)}\right)\diag(\valpha^{(t)}\odot\vy)\right\|_\mathrm{F}}_{\mathrm{Term\ 1}}\\
        &\quad + \frac{1}{4\eta}\underbrace{\left\|\diag(\valpha^{(t)}\odot\vy) \left(\mK_o\odot \mathcal{E}^{(t-1)}\right)\diag(\valpha^{(t)}\odot\vy) - \diag(\valpha^{(t-1)}\odot\vy) \left(\mK_o\odot \mathcal{E}^{(t-1)}\right)\diag(\valpha^{(t-1)}\odot\vy)\right\|_\mathrm{F}}_{\mathrm{Term\ 2}},
    \end{aligned}
\end{equation}
where inequality (a) is derived from Property~\ref{proper:1}, equality (b) originates from the definition given in equation~(\ref{equ: 7}), and inequality (c) results from applying the triangle inequality. We now first focus on Term 1.
\begin{equation}
    \begin{aligned}
        \mathrm{Term\ 1} &\leq \left\|\diag(\valpha^{(t)}\odot\vy)\|_\mathrm{F}^2 \|\mK_o\odot (\mathcal{E}^{(t)} - \mathcal{E}^{(t-1)})\right\|_\mathrm{F} \\
        &\leq NC^2 \|\mK_o\|_\mathrm{F}\| \mathcal{P}_2(\hat{\mathcal{E}}^{(t)}) -\mathcal{P}_2(\hat{\mathcal{E}}^{(t-1)})\|_\mathrm{F} \\ 
        & \overset{(a)}{\leq} NC^2 \kappa \| \hat{\mathcal{E}}^{(t)} - \hat{\mathcal{E}}^{(t-1)}\|_\mathrm{F} \\ 
        & \overset{(b)}{\leq} NC^2 \kappa \left[NC^2\kappa\left\|\mK_\Delta^{(t)}-\mK_\Delta^{(t-1)}\right\|_\mathrm{F} + 2\nu^{(t)}NC\kappa b_{\mathrm{max}}\left(1+C\kappa b_{\mathrm{max}} + \frac{NC^3\kappa\lambda b_{\mathrm{max}}^2}{4\rho}\right)\right] \\ 
        & = N^2C^4\kappa^2\left\|\mK_\Delta^{(t)}-\mK_\Delta^{(t-1)}\right\|_\mathrm{F} + 2\nu^{(t)}N^2C^3\kappa^2 b_{\mathrm{max}}\left(1+C\kappa b_{\mathrm{max}} + \frac{NC^3\kappa\lambda b_{\mathrm{max}}^2}{4\rho}\right),
    \end{aligned}
\end{equation}
where inequality (a) follows from Property~\ref{proper:1} and inequality (b) is obtained through substitution using Lemma~\ref{lemma:4}.

Subsequently, for Term 2, we can directly apply the result of Lemma~\ref{lemma:3}:
\begin{equation}
    \begin{aligned}
        \mathrm{Term\ 2} \leq 2\nu^{(t)}NC\kappa\left(1+\frac{NC^2\kappa b_{\mathrm{max}}}{4\rho}\right)\left(1+C\kappa b_{\mathrm{max}} + \frac{NC^3\kappa\lambda b_{\mathrm{max}}^2}{4\rho}\right)
    \end{aligned}
\end{equation}
Therefore, we conclude from equation~(\ref{equ:B.12}) that
\begin{equation}
    \begin{aligned}
        & \left\|\mK_\Delta^{(t+1)} - \mK_\Delta^{(t)} \right\|_\mathrm{F} \\
        & \leq \frac{1}{4\eta} N^2C^4\kappa^2\left\|\mK_\Delta^{(t)}-\mK_\Delta^{(t-1)}\right\|_\mathrm{F} + \frac{1}{2\eta}\nu^{(t)}N^2C^3\kappa^2 b_{\mathrm{max}}\left(1+C\kappa b_{\mathrm{max}} + \frac{NC^3\kappa\lambda b_{\mathrm{max}}^2}{4\rho}\right)\\
        &\quad + \frac{1}{2\eta} \nu^{(t)}NC\kappa\left(1+\frac{NC^2\kappa b_{\mathrm{max}}}{4\rho}\right)\left(1+C\kappa b_{\mathrm{max}} + \frac{NC^3\kappa\lambda b_{\mathrm{max}}^2}{4\rho}\right)
    \end{aligned}
\end{equation}
which completes the proof of the theorem.
\qed

\begin{figure}[t]
    \centering
    \includegraphics[width=0.9\linewidth]{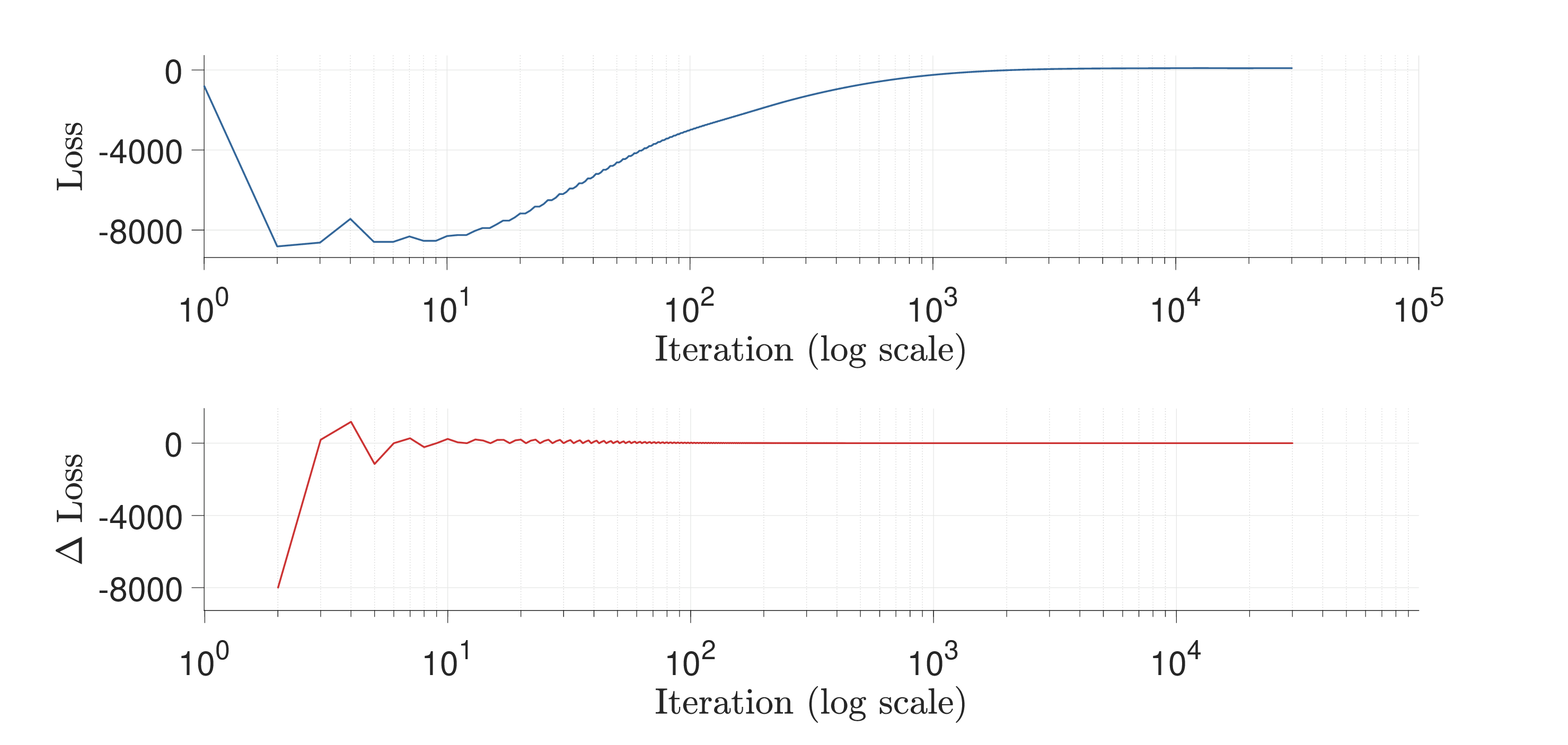}
	\caption{The convergence of the Algorithm~\ref{Algo} on the \textit{heart} dataset. Top: The loss function value changes with the number of iterations. Bottom: The difference between the loss value at iteration $i$ and iteration $i-1$.}
    \label{Fig: converge}
\end{figure}

In practice, the upper bound in Theorem~\ref{theorem:5} can be tighter. This is because, when dealing with equations such as~(\ref{equ:B.7}), we avoided overly complex analysis by not taking the dynamics of $\mK_\Delta^{(t)}$ and $\mathcal{E}^{(t)}$ into more precise consideration. Instead, we considered a static, general case $\mK_\Delta$ and $\mathcal{E}$ and directly applied the triangle inequality for simplification, resulting in this term being at the $\mathcal{O}(\sqrt{N})$ scale. On the \textit{heart} dataset, we numerically verified the convergence of Algorithm~\ref{Algo} and presented the results in Figure~\ref{Fig: converge}. The objective function value ultimately converges to 91.73 after multiple rounds of minimization and maximization. Furthermore, the difference in the objective function values between two consecutive iterations also satisfies $|\Delta\mathrm{Loss}|<1.5e^{-5}$, which supports our analysis.

\section{Four Metrics for Measuring Structural Differences in Data}
\label{secD}
The inter-class distance (ICD) measures the distance between the centroids of different classes. For two classes \( C_i \) and \( C_j \), the formula can be written as: 
\[
\text{ICD} = \| \mu_i - \mu_j \|_2,
\]
where \( \mu_i \) and \( \mu_j \) are the mean vectors (centroids) of classes \( C_i \) and \( C_j \), respectively.

Fisher discriminant ratio (FDR) evaluates the ratio of between-class variance to within-class variance: 
\[
\text{FDR} = \frac{\text{Tr}(S_b)}{\text{Tr}(S_w)},
\]
where
\[
S_b = \sum_{i=1}^k n_i (\mu_i - \mu)(\mu_i - \mu)^\top
\]
is the between-class scatter matrix, \( \mu_i \) is the mean of class \( C_i \), \( \mu \) is the overall mean, and \( n_i \) is the size of class \( C_i \). 
\[
S_w = \sum_{i=1}^k \sum_{x \in C_i} (x - \mu_i)(x - \mu_i)^\top,
\]
is the within-class scatter matrix.

Calinski-Harabasz index (CHI) measures cluster compactness and separation: 
\[
\text{CHI} = \frac{\text{Tr}(S_b)}{\text{Tr}(S_w)} \cdot \frac{N-k}{k-1},
\]
where \( S_b \) and \( S_w \) are as defined above. \( N \) is the total number of data points, and \( k \) is the number of clusters (or classes).

Davies-Bouldin index (DBI) evaluates the average similarity ratio of each cluster with the cluster most similar to it: 
\[
\text{DBI} = \frac{1}{k} \sum_{i=1}^k \max_{j \neq i} \left( \frac{\sigma_i + \sigma_j}{\| \mu_i - \mu_j \|} \right),
\]
where \( \sigma_i \) is the average intra-cluster distance for cluster \( C_i \), \( \mu_i \) is the centroid of clusters \( C_i \), and \( \| \mu_i - \mu_j \| \) is the distance between the centroids of clusters \( C_i \) and \( C_j \).

\section{Extension from Classification to Regression}
\label{secE}
To extend from the current classification model to a regression model, we can refer to the extension from support vector classification to support vector regression (SVR). The optimization objective can be modified as follows:
\begin{equation}
    \begin{aligned}
        \min_{\mK_{\Delta}}\max_{\hat{\valpha}, \check{\valpha}, \mathcal{E}}\quad& -\frac{1}{2}(\hat{\valpha}- \check{\valpha})^\top\left(\mK_o\odot\mK_{\Delta}\odot\mathcal{E}\right)(\hat{\valpha}- \check{\valpha})+(\hat{\valpha}- \check{\valpha})^\top\vy \\
        & -\varepsilon(\hat{\valpha}+ \check{\valpha})^\top\vone+\eta\|\mK_{\Delta}-\vone\vone^\top\|_{\mathrm{F}}^2\\
        \st\quad & \mB_l\preceq \mK_{\Delta} \preceq \mB_u,\  \mK_{\Delta}\in\mathcal{S}_{+},\\
        & \|\mathcal{E}-\vone\vone^\top\|_{\mathrm{F}}^2 \leq r^2,\ \mathcal{E}\in\mathcal{S}_{+},\\
        & (\hat{\valpha}- \check{\valpha})^\top\vone=0 , \ \vzero\leq \hat{\valpha}, \check{\valpha}\leq C\vone.
    \end{aligned}
    \label{Equ: SVR}
\end{equation}
For the updates of $\mK_\Delta$ and $\mathcal{E}$, the steps in the main body of this paper can be followed, while the updates for $\hat{\valpha}$ and $\check{\valpha}$ in SVR need to be adjusted. Of course, the convergence analysis of this framework would also become more challenging.







\end{document}